\documentclass[12pt,a4paper,final]{iopart}

\usepackage{graphicx}
\usepackage[breaklinks=true,colorlinks=true,linkcolor=blue,urlcolor=blue,citecolor=blue]{hyperref}

\expandafter\let\csname equation*\endcsname\relax
\expandafter\let\csname endequation*\endcsname\relax
\usepackage{amsmath}
\usepackage{amssymb}

\usepackage{multirow}
\usepackage{adjustbox}

\usepackage{subcaption,siunitx,booktabs}

\begin{document}

\title[On evolutionary selection of blackjack strategies]
{On evolutionary selection of blackjack strategies}

\author{Mikhail Goykhman}
\address{Racah Institute of Physics, Hebrew University of Jerusalem,\\
Jerusalem, 91904, Israel}
\ead{\mailto{michael.goykhman@mail.huji.ac.il} \footnote{\mailto{goykhman89@gmail.com}}}

\begin{abstract}

We apply the approach of evolutionary programming to the problem of optimization
of the blackjack basic strategy.
We demonstrate that the population of initially random blackjack strategies evolves and saturates to a profitable
performance in about one hundred generations. The resulting strategy resembles the known blackjack
basic strategies in the specifics of its prescriptions, and has a similar performance.
We also study evolution of the population of strategies initialized to the Thorp's basic strategy.

\end{abstract}

\section{Introduction}

The game of blackjack has received a considerable attention from
game theorists (and professional gamblers) during the recent decades. The original work which has
initiated such an active interest in blackjack was published
in  \cite{Baldwin1956},
and \cite{Thorp1961,Thorp1962}. In \cite{Baldwin1956} it has been
shown that following the prescribed set of rules the player can get almost even with the
dealer, reducing the casino's edge to just about 0.62\%.
\footnote{An edge (positive or negative) of $p\%$ means that for every dollar the player bets it is
expected to (make or lose) $|p|$ cents.}
Later \cite{Thorp1962} refined the edge estimate for the
strategy of \cite{Baldwin1956} to the player's advantage of $0.09\%$.
Thorp also improved
the prescription of \cite{Baldwin1956}, and claimed that, under certain specific game rules,
player following the Thorp's optimized
basic strategy has 0.12\% edge over the casino. The player's edge
can go up to $0.6\%$ under more favorable casino rules
(the edge can further increase to about 2\% if the natural pays 2:1, which we will not be
considering in this paper).
The term basic strategy implies that the player makes its decisions taking into account only its
own cards and the dealer's up-card, without taking into consideration any
other information, such as the cards which have already been played.
Such a small advantage of the basic strategy
makes the game practically even. This is essentially different from the other games
against the house,
known for a substantial player's disadvantage (such as 5.26\% disadvantage in the roulette game with
38 pockets, having one zero and one double-zero),
resulting in the certain long-term ruin for the gamblers.

Moreover, Thorp has demonstrated, see \cite{Thorp1961,Thorp1962},
that removing some cards from the deck, the player,
following the corresponding version of the basic strategy, gets the higher/smaller edge
over the house. The most significant effect is when the fives are removed from the deck,
giving player the edge of 3.58\% under the corresponding most optimal basic strategy, see \cite{Thorp1961,Thorp1962}. Therefore, as some rounds of the game are played, and the cards
are removed from the deck (in most of the blackjack games the played cards remain discarded
until the deck is exhausted or substantially depleted, in other words,
blackjack is not a game of independent trials), the player can adjust the size it
bets in accordance with the edge it expects to have on the next round of play.
Thorp has devised a counting strategy which assigns weights to the cards,
so that as the player sees the dealt cards, it adds up their total weight, thereby calculating the
count of the remaining deck. This count tells the player whether the deck 
is expected to give the player an advantage, and if yes, the player bets more.
Furthermore, the player can optimize its play by using a variant of the basic strategy
adjusted to the specific depleted deck, see \cite{Thorp1961,Thorp1962}.
Subtle issues exists as to how the player should conceal its advantage play, 
to avoid negative reaction and counter-measures from the casino employees,
which we will not be discussing in this paper, referring reader to \cite{Thorp1962}.\\

The goal of this paper is to develop and employ a numerical simulation method
for constructing and exploring optimal blackjack strategies. Simulation
methods have been routinely used for the blackjack research, including the
original work by \cite{Thorp1961,Thorp1962} (see also \cite{Millman1983}). The idea is that simulating a
sufficiently large number of hands one can estimate the edge which any
given strategy will give to the player. Typically one starts by testing a specific basic strategy,
and then extends this approach to studying the counting strategies. 
The starting point for the strategy search can be taken as the prescription of 
\cite{Baldwin1956}. One then adds some variations on top of it, and verifies whether
the resulting performance improves or not. Similarly, one develops the optimal
strategies and calculates the corresponding player's edge for the decks which
have some of the card ranks depleted (relatively to the size of the deck), or totally missing,
as described in \cite{Thorp1961,Thorp1962}.

The basic strategy optimization search therefore has two essential steps. One starts with
some known basic strategy, and, first, reasons what parts of it could be altered
in hope of obtaining a better performance. Second, one runs a simulation
in order to calculate performance of the so obtained new strategy.
If the first step is replaced with a brute force search over the parameters of
the basic strategy, the time needed to solve the problem of finding the 
optimal basic strategy will increase exponentially with the number of parameters
one desires to tweak. If one is planning to brute force search over the entire 
set of blackjack basic strategy parameter space, the search will take about $2^{620}\sim10^{186}$
steps, unfeasible for any computational resources (the origin of this estimate is discussed
in section \ref{thorp_section}).

We suggest to get around this problem by applying the ideas of evolutionary computation
to the problem of optimization of blackjack strategies. Previous work on evolutionary selection
of blackjack strategies can be found in \cite{Kendall2003}, which used evolutionary
selection to find an optimal neural network, used to analyze the blackjack game,
and \cite{Fogel2004}, which applied evolutionary selection directly to the population
of blackjack strategies
(see also \cite{Yakowitz1992}, \cite{Coleman2004} and references therein).
We propose a framework to encode the blackjack strategy into
the chromosome, and prescribe the fitness, selection, and replication rules
for the strategy chromosomes. We discuss how tuning parameters of the evolutionary
selection one can improve the evolutionary performance. It turns out that the 
evolutionary optimization search saturates well below $1000$ evolutionary steps, and works
well on the population of only $5000$ strategies,
making the problem easily solvable with the contemporary computational resources.

To search for an optimal blackjack basic strategy evolutionary we suggest two approaches.
In the first approach we start from the population of randomly initialized strategies, and let it evolve
for 1000 generations. We choose the amount of money which the strategy
generates (after $N=10^4$ rounds of play, betting $b=\$2$ per round) as its fit score, selecting and breeding the top 5\% of strategies which generate
the most money. We observe that the fit score quickly saturates in about 100 evolutionary steps.
The resulting strategy performs similarly well to the Thorp's basic strategy, and is very resembling
of it in the specifics. However, some of the detailed prescriptions, such as some
of the split and double down prescriptions, are different for the Thorp's and the 
evolved basic strategy derived in this paper.

We suggest that in the evolutionary setup of
this paper, with $N=10^4$ rounds of $b=\$2$ bet per round, some of the changes in the
strategy parameters can appear relatively non-influential.
This can result in an impression that 
several distinct blackjack basic strategies perform similarly well, and are practically
indistinguishable.
Specifically this is reflected in the fact that we cannot arrive at a statistically
significant rejection of the null hypothesis, stating that
two strategies (such as, the Thorp's and the evolved basic strategy) perform differently on average.
This artifact can be removed,
provided we use a larger number of the game rounds $N$, at expense of increasing the computational time.
In that case the difference in performance between any two distinct basic strategies
can be made as large as desired, as we demonstrate explicitly. Correspondingly, the search for the most optimal
blackjack basic strategy will be most accurate in the setup where the gradient
of the strategy fit function is made large.

To further test our approach we also looked at evolution of the population of strategies which
have been initialized identically to the Thorp's basic strategy. Similarly to what we
have done for the initially random population of strategies, we have evolved this population
of the Thorp's strategies for 1000 rounds. The resulting strategy has converged
to a slightly different mutation.
We then compared performance of the Thorp's basic strategy and the evolved
Thorp's basic strategy, by running $N=10^5$ rounds of play with $b=\$2$ bet per round.
This test demonstrated that the evolved Thorp's strategy has a statistically significantly better
performance. At the same time, running $N=10^4$ rounds of play,
is not enough to make a statistically significant statement for the difference
between the Thorp's and the evolved Thorp's strategies performance, in agreement with
the observation described above.

The optimal basic strategy depends on the specific rules which 
the casino sets.
We review the rules in section \ref{thorp_section},
and provide a brief summary here.
The rules which we follow in this paper are quite favorable for the player. Specifically, dealer stands on soft 17,
the natural pays 3:2,
money don't exchange hands at the draw (push),
player can double down on any original hand (except for 21),
player can split any hand,
player cannot double down on split aces (but can double down on other split pairs), player can hit each hand of split
aces at most once, player cannot split again the split hands,
a single deck of cards is used. For most of the basic strategy estimates we reshuffle
the deck after it has been depleted by one third. A single player plays
against the dealer in all of our simulations. Player never takes an insurance.

The rest of this paper is organized as follows. In section \ref{thorp_section}
we start by reviewing the blackjack rules.
We then discuss the Thorp's basic strategy and evaluate its performance. We look at the 
player's final bankroll after a large number of betting rounds, and discuss the
performance of the double down and split prescriptions. In section \ref{evolutionary_discussion_section}
we outline the basic principles for the evolutionary selection of strategies which we designed
and used in this paper. We also discuss the appropriate choice of the evolution parameters,
such as the selection rate (the fraction of the most fit strategies which are selected in the given
generation and are chosen for breeding), and the mutation rate of the genes.
In section \ref{section_evolving_random} we discuss evolving a population of strategies
which have been initialized randomly.
In section \ref{section_evolving_thorp} we discuss the results of evolving the population
of strategies which have been initially set to the Thorp's basic strategy.
We conclude and discuss our results in section \ref{section_conclusions}.

\section{Performance of the Thorp's basic strategy}
\label{thorp_section}

In this section we discuss our results for the performance test of 
the Thorp's basic strategy for the full deck, see \cite{Thorp1962}. We applied the methods used in this section
for evaluation of the other variants of the (evolved) basic strategies,
therefore we start with a general discussion.

The player starts with $B_0=\$1000$ bankroll.
Each round the player bets $b=\$2$. It plays the game against the dealer for
$N=10000$ rounds (some of the rounds the player splits, and plays two hands per round;
for the typical basic
strategy, such as Thorp's, the player splits about 2\% of the time;
the total number of rounds played is then about $T\simeq 10200$),
or until it goes broke. We record the number of times the 
player wins/loses/draws in total, the number of splits and double downs,
and the number of won/lost splits and double downs. This simulation is then
repeated for $M=1000$ times, allowing us to calculate distribution
properties of the recorded quantities. We can judge how successful is the
prescribed basic strategy for split/double down, and whether overall
the player ends up with a larger or smaller bankroll.

To evaluate performance of the strategy one could look at the winning
percentage $p_w=W/T$ and the losing percentage $p_l=L/T$,
where $W$ is the total number of times the player won, and $L$
is the total number of times the player lost. We notice, however, that the $p_w-p_l$ metric
does not actually reflect the true edge of the player, since it ignores the natural payoffs,
and, importantly, the winning rate of the double downs. One of the important metrics of the
strategy's performance is therefore the edge of its double downs; clearly the 
strategy should prescribe to double down only if this edge is positive for the player.
Similarly, the player's edge should be positive for the split prescriptions of the strategy.
We study these metrics for all the strategies discussed in this paper.

\subsection{Review of the blackjack rules}

In this subsection we briefly review the rules of blackjack. The specific rules, as mentioned
in Introduction section, vary between the casinos, and the corresponding optimal strategies
vary as well. The game is played by one or several players against 
the dealer. The dealer deals the cards from a single deck or a shoe, the latter is made of
several decks shuffled together. In this paper we consider the situation when only a single deck is used,
and only one player plays against the dealer.

Each card is assigned a numerical value. The number cards are assigned their corresponding
values, the face cards (jack, queen, king) are assigned the value 10, the player's ace can be either 1 or 11,
depending on the choice of the player. If the hand (the dealt set of cards) contains an ace
which can be counted as 11 without the total value of the hand going over 21, the hand is called
soft, otherwise the hand is called hard. The goal of the game is to get the total hand value below or equal to 21,
avoiding going over 21 (busting), and having that value strictly higher than the dealer's value
(or not busting, while having the dealer bust). At the showdown the hand with the highest value wins. In case
of a draw (push) no money exchanges hands.

The cards are dealt in the following order. The dealer deals two cards (typically, face up)
to the players, and two cards to itself. One of the dealer's cards is dealt face up,
another is dealt face down. Each player then faces the sequence of decisions.
If the player has two cards of the same value, it can split the hand. Variants
of the game put restrictions on what card pairs can be split (say, whether queen and ten
can be split), we will consider the rules in which any pair can be split.
If the player splits the hand, then it should put an equal bet on the table,
and then the dealer deals one more card to each of the split hands. We will
consider the rules in which the split hands cannot be re-split again (which can be
asked in case when
the split hand is dealt a card in pair).

On the original two-card hand the player can decide to double down on its bet.
Player can also double down on one or both of its split hands. We consider the game
variant in which player cannot double down if the aces were split. If player doubles
down, the dealer deals exactly one more card to the player. 

If player doesn't double down, it can request to stand, in which case it receives no more
cards, or hit, in which case the dealer deals one more card. Player can then ask to hit until
it decides to stand, or until it busts, that is, goes over 21. If aces are dealt, player can decide
to count aces as 11 or as 1. We consider the game variant in which player can hit split aces
just once.

At the showdown, provided the player hasn't busted, the dealer flips its down-card and deals cards
to itself until it goes to 17 or higher. The aces of the dealer's hand are counted as 11,
unless it causes the hand to go over 21. A variant of the game which we will not be considering 
in this paper is when the dealer hits the soft 17. Each player plays against the dealer,
and wins if its count is higher than the dealer's, or if the dealer busts. If the player busts, then
it does not matter whether the dealer would bust as well, and the round is lost for the player.
This rule is what results in the original edge for the dealer.

If the originally dealt pair of cards is of the value 21, that is, ten and ace,
it is called natural. Natural typically pays at 3:2, unless the dealer also has the
natural, which results in a draw. The pair of the value 21 on the split hands is not considered
a natural. In this paper we consider the rules variant in which the player cannot double
down on the natural (in which case the natural would have been counted as 11 instead),
the latter could be advantageous in some cases, considering that the natural pays 3:2,
and the double down win pays 2:1.

Typically in blackjack if the dealer's up-card is ace, then the first thing which the player
is allowed to do is to take insurance against the dealer's hand being a natural.
For insurance the player can put at most half of its original bet, and if the dealer has
a natural (which it then checks right away, without revealing the down-card value
if the hand is not a natural), then the insurance is paid at 2:1. Otherwise the dealer collects
the insurance bet and the round continues. In this paper we consider strategies which do not take insurance.
This is because in situations when only the player's hand cards are taken into account
(as it is in the basic strategy framework, as contrasted with the card counting approach),
the insurance is at $2\%$-$14.3\%$ disadvantage to the player, see \cite{Thorp1962}.

\subsection{Thorp's basic strategy}

In this section we report the results of the simulation for the Thorp's
basic strategy, see tables \ref{thorp_split_soft_hard_dd},  \ref{thorp_stand}.
For our purposes it is convenient to represent the strategy prescriptions
in terms of tables filled with ones and zeros, where one indicates that the
player should split/double down/stand, and zero indicates that the
player shouldn't split/shouldn't double down/should hit. Notice that some
of the table entries are redundant, for instance, in table \ref{thorp_split_soft_hard_dd}
for the soft double down the player will never double down on (A,T),
and for the hard double down the player will never double down
on the count of 21. Similarly, in table \ref{thorp_stand} for the soft stand the
counts of 2-11 are redundant, since every soft hand has one ace counting as 11,
making the total score larger than 11. The last rows in the soft/hard stand
tables \ref{thorp_stand} are also redundant, because the player always stands on 21.
The first two rows of the hard stand and the hard double down tables are redundant,
because the hard count on the original hand of two cards
cannot be $2$ and $3$. This leaves 620 basic strategy parameters, reducing the
original count of 800 parameters.

Some other entries of the strategy tables \ref{thorp_split_soft_hard_dd},  \ref{thorp_stand},
while not redundant in general, are redundant for the specifically Thorp's basic strategy.
For instance, since the player is always prescribed to split aces, and due to the fact that
splitting decision has the priority over doubling down and standing decisions, the first row in
the soft double down table \ref{thorp_split_soft_hard_dd} is redundant. (Unless it is being
applied to the split hand, which we cannot re-split again; however according to the rules
considered in this paper the player cannot double down on split aces.)

Thorp has pointed out that the
hard double down prescription for the player's count of 8 is to be voided if player holds (6,2);
and player should stand on hard 16, against dealer's 10, if holding three or more cards. In case of the Thorp's basic strategy the performance change when
these subtleties are taken into account is discussed below.

\begin{table}[!htb]
\begin{minipage}{.5\linewidth}
\begin{subtable}{1\textwidth}
\caption*{Thorp's basic strategy prescription for split.}    
\scalebox{0.8}{
\begin{tabular}{|*{12}{c|}}
\cline{3-12}
\multicolumn{2}{c|}{\multirow{2}{*}{}} & \multicolumn{10}{c|}{Dealer's up-card}\\\cline{3-12}
\multicolumn{2}{c|}{}
& {\bf A} & {\bf 2} & {\bf 3} & {\bf 4} & {\bf 5} & {\bf 6} & {\bf 7} & {\bf 8} & {\bf 9} & {\bf T} \\\hline
\multirow{10}{*}{\rotatebox{90}{Player's paired card}}
& {\bf A} & 1 & 1 & 1 & 1 & 1 & 1 & 1 & 1 & 1 & 1 \\\cline{2-12}
& {\bf 2} & 0 & 1 & 1 & 1 & 1 & 1 & 1 & 0 & 0 & 0\\\cline{2-12}
& {\bf 3} & 0 & 1 & 1 & 1 & 1 & 1 & 1 & 0 & 0 & 0\\\cline{2-12}
& {\bf 4} & 0 & 0 & 0 & 0 & 1 & 0 & 0 & 0 & 0 & 0\\\cline{2-12}
& {\bf 5} & 0 & 0 & 0 & 0 & 0 & 0 & 0 & 0 & 0 & 0\\\cline{2-12}
& {\bf 6} & 0 & 1 & 1 & 1 & 1 & 1 & 1 & 0 & 0 & 0\\\cline{2-12}
& {\bf 7} & 0 & 1 & 1 & 1 & 1 & 1 & 1 & 1 & 0 & 0\\\cline{2-12}
& {\bf 8} & 1 & 1 & 1 & 1 & 1 & 1 & 1 & 1 & 1 & 1\\\cline{2-12}
& {\bf 9} & 0 & 1 & 1 & 1 & 1 & 1 & 0 & 1 & 1 & 0\\\cline{2-12}
& {\bf T} & 0 & 0 & 0 & 0 & 0 & 0 & 0 & 0 & 0 & 0\\\hline
\end{tabular}
}
\end{subtable}
\begin{subtable}{1\textwidth}
\caption*{Thorp's basic strategy prescription for soft double down.}
\scalebox{0.8}{
\begin{tabular}{|*{12}{c|}}
\cline{3-12}
\multicolumn{2}{c|}{\multirow{2}{*}{}} & \multicolumn{10}{c|}{Dealer's up-card}\\\cline{3-12}
\multicolumn{2}{c|}{}
& {\bf A} & {\bf 2} & {\bf 3} & {\bf 4} & {\bf 5} & {\bf 6} & {\bf 7} & {\bf 8} & {\bf 9} & {\bf T} \\\hline
\multirow{10}{*}{\rotatebox{90}{Player's other card}}
& {\bf A} & 0 & 0 & 0 & 0 & 1 & 1 & 0 & 0 & 0 & 0 \\\cline{2-12}
& {\bf 2} & 0 & 0 & 0 & 1 & 1 & 1 & 0 & 0 & 0 & 0\\\cline{2-12}
& {\bf 3} & 0 & 0 & 0 & 1 & 1 & 1 & 0 & 0 & 0 & 0\\\cline{2-12}
& {\bf 4} & 0 & 0 & 0 & 1 & 1 & 1 & 0 & 0 & 0 & 0\\\cline{2-12}
& {\bf 5} & 0 & 0 & 0 & 1 & 1 & 1 & 0 & 0 & 0 & 0\\\cline{2-12}
& {\bf 6} & 0 & 1 & 1 & 1 & 1 & 1 & 0 & 0 & 0 & 0\\\cline{2-12}
& {\bf 7} & 0 & 0 & 1 & 1 & 1 & 1 & 0 & 0 & 0 & 0\\\cline{2-12}
& {\bf 8} & 0 & 0 & 0 & 0 & 0 & 0 & 0 & 0 & 0 & 0\\\cline{2-12}
& {\bf 9} & 0 & 0 & 0 & 0 & 0 & 0 & 0 & 0 & 0 & 0\\\cline{2-12}
& {\bf T} & 0 & 0 & 0 & 0 & 0 & 0 & 0 & 0 & 0 & 0\\\hline
\end{tabular}
}
\end{subtable}
\end{minipage}
\begin{minipage}{.5\linewidth}
\caption{Thorp's basic strategy prescription for split and soft double down ({\bf left}), and for hard double down ({\bf right}).}
\label{thorp_split_soft_hard_dd}
\centering
\scalebox{0.87}{
\begin{tabular}{|*{12}{c|}}
\cline{3-12}
\multicolumn{2}{c|}{\multirow{2}{*}{}} & \multicolumn{10}{c|}{Dealer's up-card}\\\cline{3-12}
\multicolumn{2}{c|}{} & {\bf A} & {\bf 2} & {\bf 3} & {\bf 4} & {\bf 5} & {\bf 6} & {\bf 7} & {\bf 8} & {\bf 9} & {\bf T} \\\hline
\multirow{20}{*}{\rotatebox{90}{Player's count}}
& {\bf 2} & 0 & 0 & 0 & 0 & 0 & 0 & 0 & 0 & 0 & 0 \\\cline{2-12}
& {\bf 3} & 0 & 0 & 0 & 0 & 0 & 0 & 0 & 0 & 0 & 0\\\cline{2-12}
& {\bf 4} & 0 & 0 & 0 & 0 & 0 & 0 & 0 & 0 & 0 & 0\\\cline{2-12}
& {\bf 5} & 0 & 0 & 0 & 0 & 0 & 0 & 0 & 0 & 0 & 0\\\cline{2-12}
& {\bf 6} & 0 & 0 & 0 & 0 & 0 & 0 & 0 & 0 & 0 & 0\\\cline{2-12}
& {\bf 7} & 0 & 0 & 0 & 0 & 0 & 0 & 0 & 0 & 0 & 0\\\cline{2-12}
& {\bf 8} & 0 & 0 & 0 & 0 & 1 & 1 & 0 & 0 & 0 & 0\\\cline{2-12}
& {\bf 9} & 0 & 1 & 1 & 1 & 1 & 1 & 0 & 0 & 0 & 0\\\cline{2-12}
& {\bf 10} & 0 & 1 & 1 & 1 & 1 & 1 & 1 & 1 & 1 & 0\\\cline{2-12}
& {\bf 11} & 1 & 1 & 1 & 1 & 1 & 1 & 1 & 1 & 1 & 1\\\cline{2-12}
& {\bf 12} & 0 & 0 & 0 & 0 & 0 & 0 & 0 & 0 & 0 & 0\\\cline{2-12}
& {\bf 13} & 0 & 0 & 0 & 0 & 0 & 0 & 0 & 0 & 0 & 0\\\cline{2-12}
& {\bf 14} & 0 & 0 & 0 & 0 & 0 & 0 & 0 & 0 & 0 & 0\\\cline{2-12}
& {\bf 15} & 0 & 0 & 0 & 0 & 0 & 0 & 0 & 0 & 0 & 0\\\cline{2-12}
& {\bf 16} & 0 & 0 & 0 & 0 & 0 & 0 & 0 & 0 & 0 & 0\\\cline{2-12}
& {\bf 17} & 0 & 0 & 0 & 0 & 0 & 0 & 0 & 0 & 0 & 0\\\cline{2-12}
& {\bf 18} & 0 & 0 & 0 & 0 & 0 & 0 & 0 & 0 & 0 & 0\\\cline{2-12}
& {\bf 19} & 0 & 0 & 0 & 0 & 0 & 0 & 0 & 0 & 0 & 0\\\cline{2-12}
& {\bf 20} & 0 & 0 & 0 & 0 & 0 & 0 & 0 & 0 & 0 & 0\\\cline{2-12}
& {\bf 21} & 0 & 0 & 0 & 0 & 0 & 0 & 0 & 0 & 0 & 0\\\hline
\end{tabular}
}
\end{minipage}
\end{table}

\begin{table}[!htb]  
\caption{Thorp's basic strategy prescription for soft stand ({\bf left})
and hard stand ({\bf right}).}
\begin{minipage}{.5\linewidth}
\centering
\scalebox{0.8}{
\begin{tabular}{|*{12}{c|}}
\cline{3-12}
\multicolumn{2}{c|}{\multirow{2}{*}{}} & \multicolumn{10}{c|}{Dealer's up-card}\\\cline{3-12}
\multicolumn{2}{c|}{} & {\bf A} & {\bf 2} & {\bf 3} & {\bf 4} & {\bf 5} & {\bf 6} & {\bf 7} & {\bf 8} & {\bf 9} & {\bf T} \\\hline
\multirow{20}{*}{\rotatebox{90}{Player's count}}
& {\bf 2} & 0 & 0 & 0 & 0 & 0 & 0 & 0 & 0 & 0 & 0 \\\cline{2-12}
& {\bf 3} & 0 & 0 & 0 & 0 & 0 & 0 & 0 & 0 & 0 & 0\\\cline{2-12}
& {\bf 4} & 0 & 0 & 0 & 0 & 0 & 0 & 0 & 0 & 0 & 0\\\cline{2-12}
& {\bf 5} & 0 & 0 & 0 & 0 & 0 & 0 & 0 & 0 & 0 & 0\\\cline{2-12}
& {\bf 6} & 0 & 0 & 0 & 0 & 0 & 0 & 0 & 0 & 0 & 0\\\cline{2-12}
& {\bf 7} & 0 & 0 & 0 & 0 & 0 & 0 & 0 & 0 & 0 & 0\\\cline{2-12}
& {\bf 8} & 0 & 0 & 0 & 0 & 0 & 0 & 0 & 0 & 0 & 0\\\cline{2-12}
& {\bf 9} & 0 & 0 & 0 & 0 & 0 & 0 & 0 & 0 & 0 & 0\\\cline{2-12}
& {\bf 10} & 0 & 0 & 0 & 0 & 0 & 0 & 0 & 0 & 0 & 0\\\cline{2-12}
& {\bf 11} & 0 & 0 & 0 & 0 & 0 & 0 & 0 & 0 & 0 & 0\\\cline{2-12}
& {\bf 12} & 0 & 0 & 0 & 0 & 0 & 0 & 0 & 0 & 0 & 0\\\cline{2-12}
& {\bf 13} & 0 & 0 & 0 & 0 & 0 & 0 & 0 & 0 & 0 & 0\\\cline{2-12}
& {\bf 14} & 0 & 0 & 0 & 0 & 0 & 0 & 0 & 0 & 0 & 0\\\cline{2-12}
& {\bf 15} & 0 & 0 & 0 & 0 & 0 & 0 & 0 & 0 & 0 & 0\\\cline{2-12}
& {\bf 16} & 0 & 0 & 0 & 0 & 0 & 0 & 0 & 0 & 0 & 0\\\cline{2-12}
& {\bf 17} & 0 & 0 & 0 & 0 & 0 & 0 & 0 & 0 & 0 & 0\\\cline{2-12}
& {\bf 18} & 1 & 1 & 1 & 1 & 1 & 1 & 1 & 1 & 0 & 0\\\cline{2-12}
& {\bf 19} & 1 & 1 & 1 & 1 & 1 & 1 & 1 & 1 & 1 & 1\\\cline{2-12}
& {\bf 20} & 1 & 1 & 1 & 1 & 1 & 1 & 1 & 1 & 1 & 1\\\cline{2-12}
& {\bf 21} & 1 & 1 & 1 & 1 & 1 & 1 & 1 & 1 & 1 & 1\\\hline
\end{tabular}
}
\end{minipage} 
\begin{minipage}{.5\linewidth}
\label{thorp_stand}
\centering
\scalebox{0.8}{
\begin{tabular}{|*{12}{c|}}
\cline{3-12}
\multicolumn{2}{c|}{\multirow{2}{*}{}} & \multicolumn{10}{c|}{Dealer's up-card}\\\cline{3-12}
\multicolumn{2}{c|}{} & {\bf A} & {\bf 2} & {\bf 3} & {\bf 4} & {\bf 5} & {\bf 6} & {\bf 7} & {\bf 8} & {\bf 9} & {\bf T} \\\hline
\multirow{20}{*}{\rotatebox{90}{Player's count}}
& {\bf 2} & 0 & 0 & 0 & 0 & 0 & 0 & 0 & 0 & 0 & 0\\\cline{2-12}
& {\bf 3} & 0 & 0 & 0 & 0 & 0 & 0 & 0 & 0 & 0 & 0\\\cline{2-12}
& {\bf 4} & 0 & 0 & 0 & 0 & 0 & 0 & 0 & 0 & 0 & 0\\\cline{2-12}
& {\bf 5} & 0 & 0 & 0 & 0 & 0 & 0 & 0 & 0 & 0 & 0\\\cline{2-12}
& {\bf 6} & 0 & 0 & 0 & 0 & 0 & 0 & 0 & 0 & 0 & 0\\\cline{2-12}
& {\bf 7} & 0 & 0 & 0 & 0 & 0 & 0 & 0 & 0 & 0 & 0\\\cline{2-12}
& {\bf 8} & 0 & 0 & 0 & 0 & 0 & 0 & 0 & 0 & 0 & 0\\\cline{2-12}
& {\bf 9} & 0 & 0 & 0 & 0 & 0 & 0 & 0 & 0 & 0 & 0\\\cline{2-12}
& {\bf 10} & 0 & 0 & 0 & 0 & 0 & 0 & 0 & 0 & 0 & 0\\\cline{2-12}
& {\bf 11} & 0 & 0 & 0 & 0 & 0 & 0 & 0 & 0 & 0 & 0\\\cline{2-12}
& {\bf 12} & 0 & 0 & 0 & 1 & 1 & 1 & 0 & 0 & 0 & 0\\\cline{2-12}
& {\bf 13} & 0 & 1 & 1 & 1 & 1 & 1 & 0 & 0 & 0 & 0\\\cline{2-12}
& {\bf 14} & 0 & 1 & 1 & 1 & 1 & 1 & 0 & 0 & 0 & 0\\\cline{2-12}
& {\bf 15} & 0 & 1 & 1 & 1 & 1 & 1 & 0 & 0 & 0 & 0\\\cline{2-12}
& {\bf 16} & 0 & 1 & 1 & 1 & 1 & 1 & 0 & 0 & 0 & 0\\\cline{2-12}
& {\bf 17} & 1 & 1 & 1 & 1 & 1 & 1 & 1 & 1 & 1 & 1\\\cline{2-12}
& {\bf 18} & 1 & 1 & 1 & 1 & 1 & 1 & 1 & 1 & 1 & 1\\\cline{2-12}
& {\bf 19} & 1 & 1 & 1 & 1 & 1 & 1 & 1 & 1 & 1 & 1\\\cline{2-12}
& {\bf 20} & 1 & 1 & 1 & 1 & 1 & 1 & 1 & 1 & 1 & 1\\\cline{2-12}
& {\bf 21} & 1 & 1 & 1 & 1 & 1 & 1 & 1 & 1 & 1 & 1\\\hline
\end{tabular}
}
\end{minipage}
\end{table}

\subsection{Performance results}

\begin{figure}
\begin{center}
\includegraphics[width=6cm, height=3.75cm]{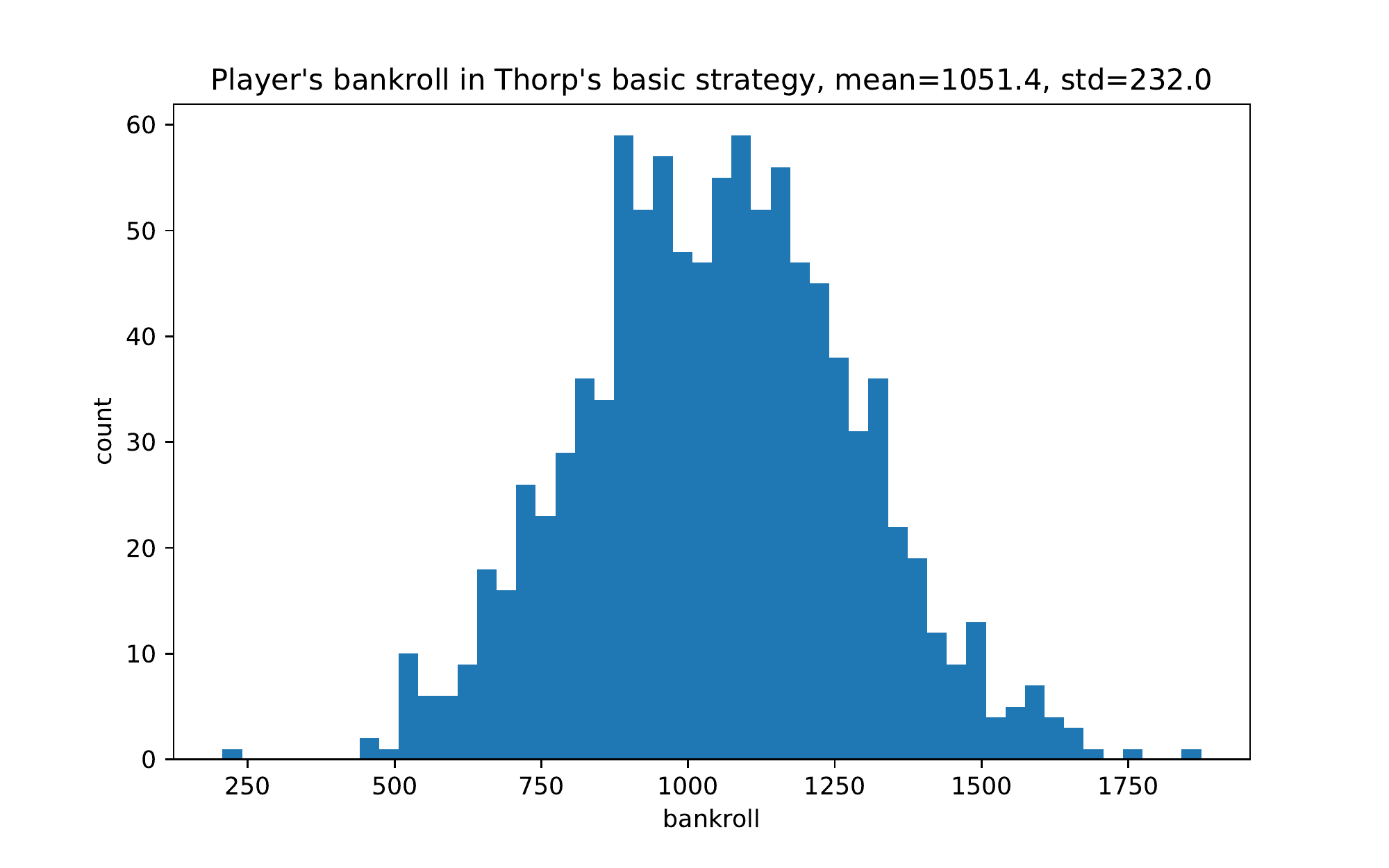}
 \includegraphics[width=6cm, height=3.75cm]{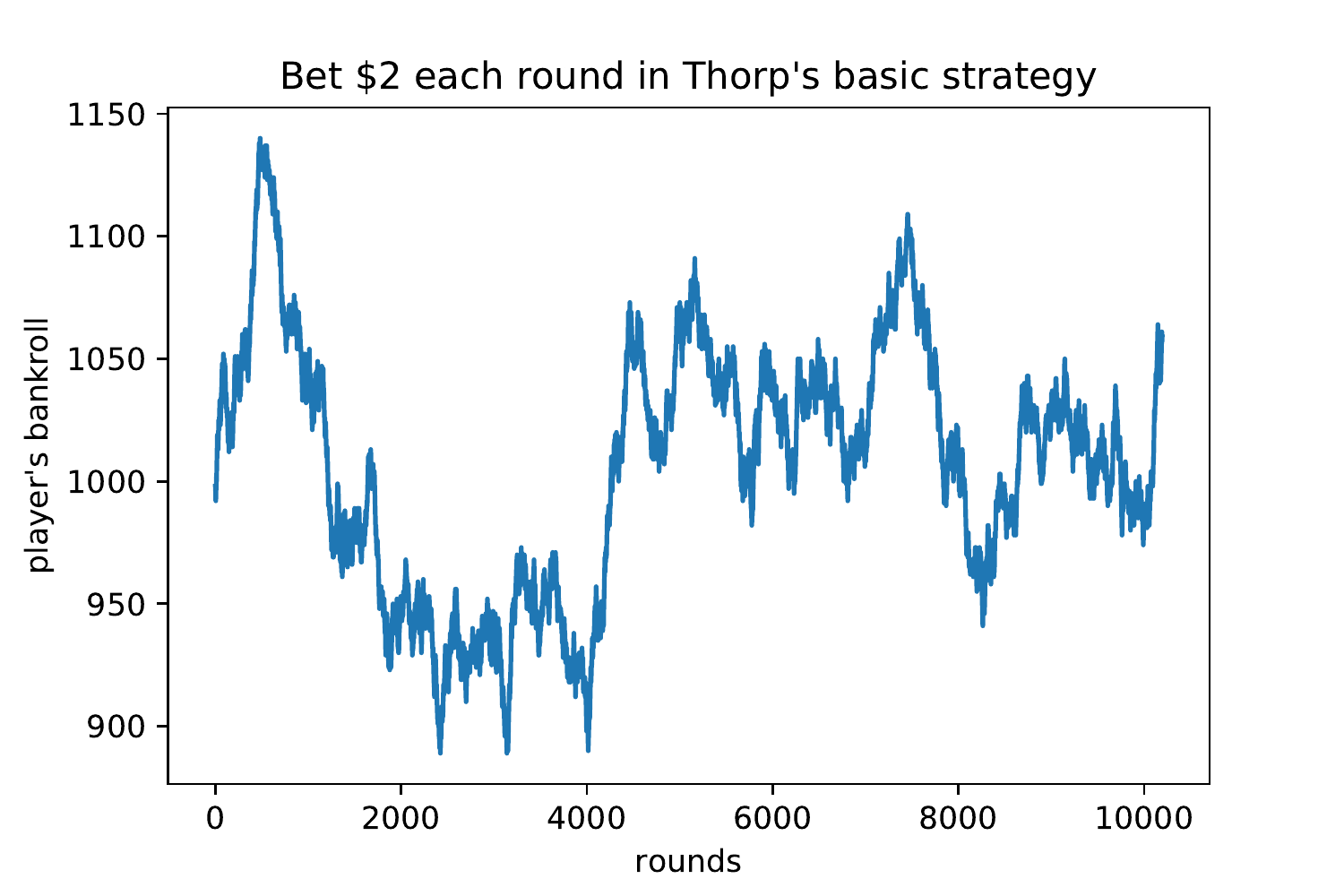}
 \caption{ \label{thorp_player_bankroll} 
 {\bf Left:} Player's final bankroll after 10000 rounds of game (with about 200 extra split rounds),
 {\bf Right:} Player's bankroll time series in one of the games; played according to Thorp's
 basic strategy (see tables \ref{thorp_split_soft_hard_dd},  \ref{thorp_stand}) in the simulation described in section~\ref{thorp_section}.
 Player starts with \$1000 bankroll ands bets \$2 per round.
 }
\end{center}
\end{figure}

In figure \ref{thorp_player_bankroll}
we plot the player's final bankroll distribution, and the sample player's bankroll
time series over the course of the game. The mean final bankroll is
$B_T=\$1051.4$, and the standard deviation is $s=\$ 232.0$, indicating the player's edge
\begin{equation}
p_e=\frac{B_T-B_0}{b_{tot}}\simeq 0.0023\,,
\end{equation}
that is, 0.23\%, where $b_{tot}=(0.11\times 2b+0.89\times b)T$
is the total money bet (in action), where probability of double down is $0.11$,
and $T\simeq (1+0.02)N$ is the total number of rounds played,
where $0.02$ is the probability of split.
The $95\%$ confidence interval for the player's
final bankroll is $[\$1037.0, \$1065.8]$, and the corresponding
$95\%$ confidence interval for the player's edge is $[0.16\%,0.29\%]$.
The fact that the Thorp's estimate for the player's edge is 0.12\%,
outside the 95\% confidence interval, probably indicates
that Thorp has used less favorable rules for the player.

We repeated this calculation of player's final bankroll, as described above,
over $100$ simulations, collecting 100 means of the final bankroll of $M=1000$ games
(each game is $T\simeq 10200$ rounds). The mean of the means of the final bankroll is $\$ 1058.2$, and the standard
deviation is $\$ 6.9$. The corresponding $95\%$ confidence interval for the mean of $M=1000$ of the Thorp's
basic strategy performance outcomes is $[\$ 1056.8,\$ 1059.6]$. This indicates the $95\%$ confidence
interval for the player's edge $[0.25\%,0.26\%]$.

\begin{figure}
\begin{center}
\includegraphics[width=6cm, height=3.75cm]{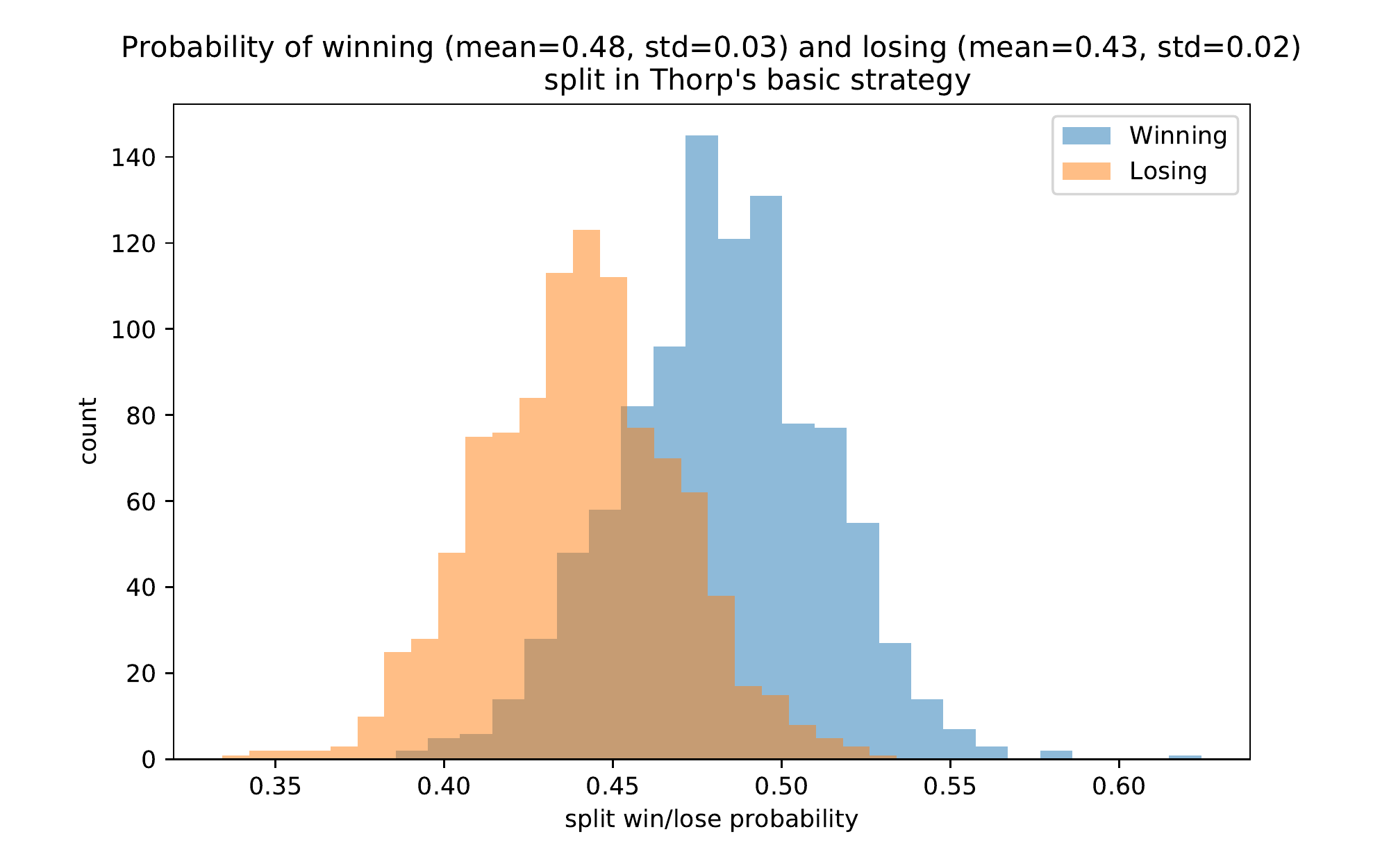}
 \includegraphics[width=6cm, height=3.75cm]{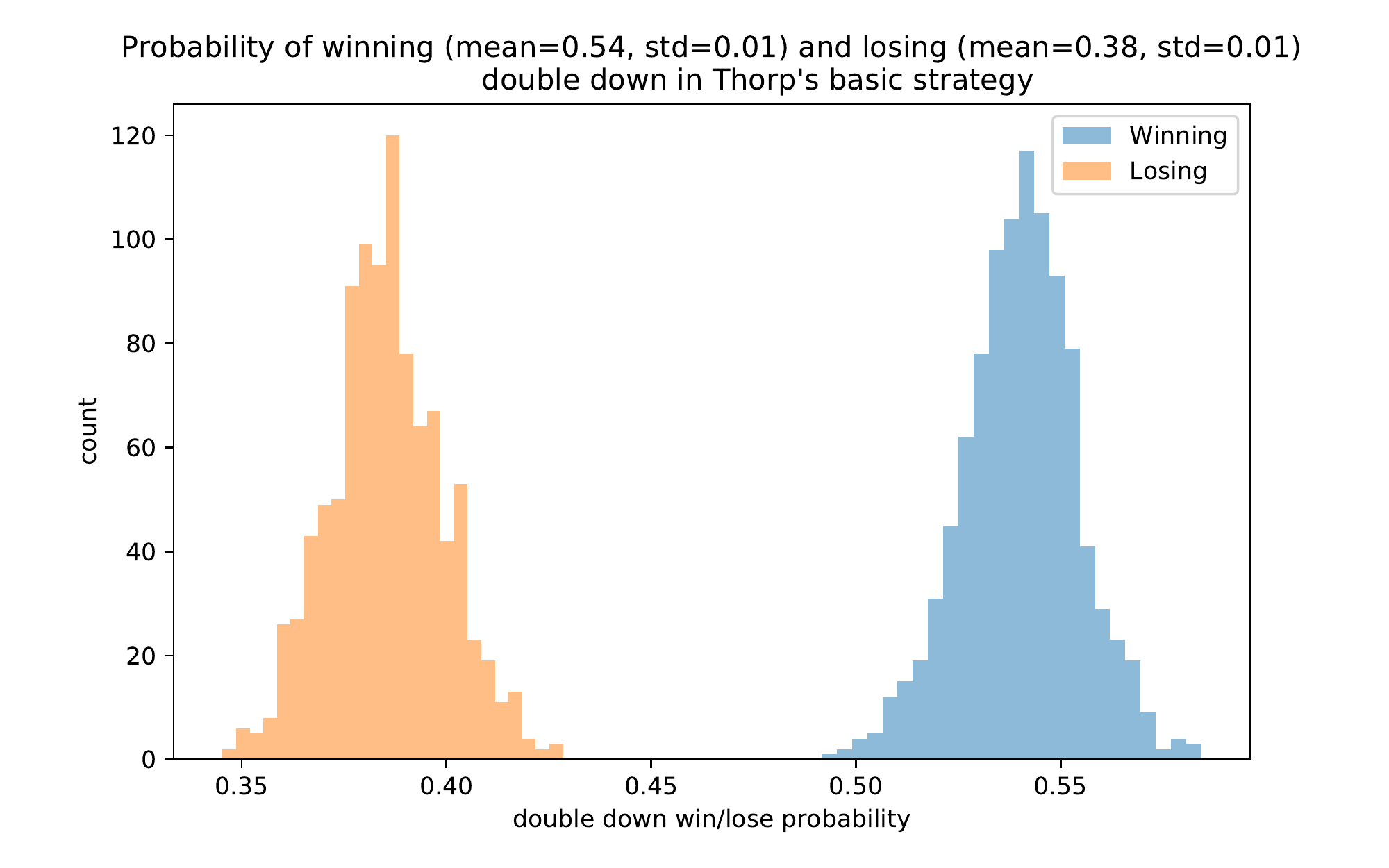}
 \caption{ \label{thorp_split_dd_win_loss} 
 Probability of winning/losing split ({\bf left}) and double down ({\bf right}) in Thorp's basic strategy
 (see tables \ref{thorp_split_soft_hard_dd},  \ref{thorp_stand}) in the simulation described in section~\ref{thorp_section}.
 The mean probability of split is 0.02, the mean probability of double down is 0.11
 }
\end{center}
\end{figure}

We also calculated that for the Thorp's strategy the player's win probability
minus the player's loss probability is $p_w-p_l\simeq -0.018$
(where the mean overall win and loss probabilities are $p_w=0.446$,
$p_l=0.464$),
which indicates the $1.8\%$ casino advantage. As pointed out above in this section,
such a conclusion would be premature, since it assumes that all the rounds
win or lose equally, thereby ignoring the different payoffs of the blackjack natural,
and the advantage of an optimal double down play.
The important metrics of the strategy performance are the edge of the split and the 
double down play, which we present in figure \ref{thorp_split_dd_win_loss}.

Taking into account the refinements of the Thorp's basic strategy, prescribing to stand on hard 16,
against dealer's 10, when holding 3 or more cards,
and not to double down on (6,2), changes mean player's bankroll
to  $\$ 1055.1$, and standard deviation to $\$ 233.3$.
We repeated this calculation of player's final bankroll, as described above,
over $100$ simulations. The mean final bankroll is $\$ 1057.4$, and the standard
deviation is $\$ 7.3$. The corresponding $95\%$ confidence interval
is $[0.25\%,0.26\%]$, practically indistinguishable in the current setting from the result above, which ignored these
extra refinements.

\section{Blackjack strategy chromosome and evolutionary selection}
\label{evolutionary_discussion_section}

Evolutionary programming is a powerful approach to optimization problems for systems with 
large number of parameters.
In case of the blackjack basic strategy optimization problem
each parameter takes one of the two values (0 or 1), prescribing
for the player how to act in each particular game situation.
\footnote{This specific parametrization of the problem ignores the subtle distinctions, such as, {\it e.g.},
what specific cards make the given total hand value.}
This parametrization of the
blackjack basic strategy can be illustrated by the Thorp's basic strategy, discussed in section \ref{thorp_section},
see tables \ref{thorp_split_soft_hard_dd},  \ref{thorp_stand}, which we can serialize into the vector $v$,
\begin{equation}
\label{splice_chromosome}
v=[v_{{\rm split}},\, v_{{\rm soft\, dd}},\,v_{{\rm hard \, dd}},\,v_{{\rm soft\, stand}},\,v_{{\rm hard\, stand}}]\,,
\end{equation}
of the dimension $d=100+100+200+200+200=800$.
We choose each matrix to be serialized row-by-row.
The brute force search over these parameters would take $2^d \simeq 10^{240}$ steps, making it unfeasible for
any reasonable computational power.

It is therefore natural to try to tackle the problem of finding the optimal (or close to the optimal)
blackjack strategy by the evolutionary programming approach.
In this section we describe our evolutionary selection method, applied to the population of the
blackjack strategy chromosomes. We will use the evolutionary algorithm described in this section 
to evolve the population of initially random blackjack strategies in section~\ref{section_evolving_random},
and the population of strategies identically initialized to the Thorp's basic strategy in section~\ref{section_evolving_thorp}.

In the evolutionary programming the strategy vector (\ref{splice_chromosome}) takes the role of the chromosome,
and its entries are assigned the role of genes. Some of the genes have a phenotypic expression by indicating to the player
how to act in the corresponding game situation. In the specific parametrization (\ref{splice_chromosome}) some
of the genes have no phenotypic expression, as discussed in section \ref{thorp_section}. For instance, the first ten rows of the soft stand matrix (encoded
in the first 100 values of its serialized representation $v_{{\rm soft\, stand}}$) are never used to make a decision
during the game. This is because they correspond to the player's count from  2 to 11 in the soft hand, while the soft hand's count
is always larger than 11. 

Some of the
genes, even under an optimized chromosome design, might end up with a suppressed phenotypic expression.
For instance, if the basic strategy prescribes to double down on the hard count of 11, then it does not matter what the
hard stand genes are for the count of 11 (when applied to the count of 11 on the originally dealt hand of two cards) because the decision to double down is taken before
the decision to stand. It is interesting to consider the chromosome choice where some of the
genes are superfluous for the game decision making, and see what values those genes will have in 
evolved chromosomes. We will see that in the most cases an evolved population will have
some genes with no phenotypic expression having the values of 1 or 0, almost identically for the entire population.
This can be explained by noticing that some of those genes might have been selected at an earlier stages of evolution,
and then have persisted throughout the rest of the evolution by just being successful at replicating themselves.
Indeed, if the population has, for instance, the given gene being predominantly equal to 1, it is natural
to expect that the winning chromosomes will also happen to have that gene being equal to 1, and have that gene more
likely to reproduce and propagate as 1.

 We need to decide how we are going to rate each strategy chromosome $v$.
In the evolutionary computation terms we need to decide on the fitness function $f(v)$,
which will return a score number for each particular strategy.
To calculate a fit value for the strategy we need to simulate a game using that strategy.
In this paper we chose to play the game with a single player against the dealer for $N=10^4$
rounds. Some of the rounds will be split, which for the evolved
strategies amounts to adding about 2\% more games, making the total
number of games played about $T\simeq 10200$.


We select the players ending up with a higher final bankroll $B_T$.
More precisely, we chose the return on the player's bankroll,
\begin{equation}
\label{fit_rank}
\phi=\frac{B_T-B_0}{B_0},
\end{equation}
as the fit score $f(v)$, where all the players start with the same bankroll $B_0=\$ 10^4$ (this choice
of the initial bankroll of the evolving strategies is an order of magnitude higher than the initial bankroll
$\$ 10^3$ chosen in the tests
of the optimal and evolved strategies, because the strategies in the earlier stages of evolution will lose a lot of money), and bet $b=\$ 2$
per round. The most fit strategies will reasonably be chosen as the strategies which make the most money.
We let the specific ways in which the strategy achieves this goal be implicit, and optimized over the
course of evolutionary selection.

The evolutionary selection happens according to the following algorithm.
We initialize a population of $M=5000$ agents, each endowed with a chromosome
$v_i$, $i=1,\dots,M$. This chromosome $v_i$ is a vector of the length $d=800$, each of its
entries (genes) is either $0$ or $1$. The phenotypic expression of the chromosome is determined
by first parsing it into five vectors (\ref{splice_chromosome}), and then
de-serializing (we have chosen the row-by-row convention) each vector into a matrix.
Those matrices tell the player how to act, where the value of $1$
of the matrix entry gives the answer `yes', and $0$ gives the answer `no',
to the question of whether the player should split/double down/stand  in the corresponding game situation. This is illustrated by tables
\ref{thorp_split_soft_hard_dd},  \ref{thorp_stand}, on example of the Thorp's basic strategy,
for instance, the table \ref{thorp_split_soft_hard_dd} indicates that the player should double down
on the soft hand (A,6) against the dealer's up-card being 2.

The initial population of strategies can be initialized randomly, or to some specific values.
It is interesting to consider a population of chromosomes, each initialized without any strategy
preconception, assigning to each of its genes a value of 0 or 1 with 50-50\% probability.
We will look at evolution of such a system in section \ref{section_evolving_random}.
In section \ref{section_evolving_thorp} we attempt to optimize the Thorp's basic strategy,
and study the question of its evolutionary stability.
To this end, we will initialize a population of the Thorp's strategies, and allow it to evolve.

Each strategy $v_i$ plays the game against the dealer for $N=10^4$ rounds
(the actual number of rounds played will be $T\simeq 10200$,
due to some split rounds), starting
from the same bankroll $B_0=\$ 10^4$, and betting $b=\$ 2$ per round. The strategies are then
sorted according to their performance, using the score (\ref{fit_rank}), reflecting their money finishes.
We select the fraction
\begin{equation}
\label{alpha_rate}
\alpha=0.05
\end{equation}
of the most fit strategies. The selection rate $\alpha$ should be small
enough to select the most optimal strategies and avoid noise coming from un-fit chromosome mutations, but not too small, so as to avoid overfitting
(such as, picking a small statistically insignificant number of `lucky' strategies).
The $1-\alpha$ of the least fit strategies are discarded from the population.

We then breed the selected $\alpha M$ strategies, until they replenish the population to the size $M$,
by producing $(1-\alpha) M$ offsprings. Each offspring is produced from two parents.
To achieve this we select two different parents $i$, $j$, with the probabilities proportional to their fit scores,
\begin{equation}
\label{probability_parent}
p_i=\frac{\phi_i}{\sum _{j=1}^{\alpha M}\phi_j}\,.
\end{equation}
This way the more fit strategies will be selected for breeding more frequently,
which is a natural setting.

Once two parent chromosomes $v_i$, $v_j$, are chosen for breeding, they produce one offspring
chromosome $c$ according to the following rule. Each gene $c^a$, $a=1,\dots, d$
of the offspring chromosome gets its value ($0$ or $1$) depending on the value of the corresponding
genes $v_i^a$, $v_j^a$ of the parents's chromosomes. If $v_i^a=v_j^a$, then $c^a$ is assigned
this same value, $c^a=v_i^a=v_j^a$, with the probability
\begin{equation}
\label{propagation_rate}
\pi=1-10^{-4}\,.
\end{equation}
With the probability $1-\pi=10^{-4}$
the $c^a$ will mutate, getting assigned the flipped value, $c^a=(1+v_i^a)\%2$.
At the mutation rate (\ref{propagation_rate}) each gene in the population of $M=5000$
chromosomes will take two generations on average to mutate. In the setup of section \ref{section_evolving_thorp},
where we start with the population of identical chromosomes, the second
generation will have about $Md(1-\pi)=400$ mutated genes.

Increasing the mutation probability $1-\pi$ speeds up the evolution, but makes it saturate to a sub-optimal value.
We want $1-\pi$ to be large enough to have an evolution happening relatively fast, but small enough to
prevent destruction of the good chromosomes and avoid convergence of the evolution to a sub-optimal performance.

If the parents have a different genes at the slot $a$, $v_i^a\neq v_j^a$, then the child's gene $c^a$
is assigned the value of one of its parents's genes, with the probabilities determined
by the fit scores.\footnote{Unlike selection of the parents for breeding with the
probabilities (\ref{probability_parent}) proportional to their fit scores,
selecting a gene over its allele with the probability proportional to its fit score
is an upgrade compared to the natural selection. In nature it is not reasonable to expect
that an advantageous phenotypic expression of the gene also coincides
with it being more likely to be chosen in the meiosis, and dominate over the corresponding gene of the
other parent.}

We select pairs of parents from the original $\alpha M$ of the most fit strategies for the total of $(1-\alpha )M$
times, replenishing the total count of the population to $M$. For the current generation we save the mean
score
\begin{equation}
\label{mean_score_most_fit}
\hat\phi=\frac{1}{\alpha M}\sum_{i=1}^{\alpha M}\phi_{(i)}\,,
\end{equation}
 of the $\alpha M$ most fit strategies. From the 
the dependence of the score $\hat\phi(t)$ on the number of the evolution steps $t$ we can observe convergence
of the evolutionary selection. The evolution should be run until at least the score (\ref{mean_score_most_fit})
saturates.

\section{Evolving random strategies}
\label{section_evolving_random}

\begin{figure}
\begin{center}
 \includegraphics[width=12cm,height=12cm,keepaspectratio]{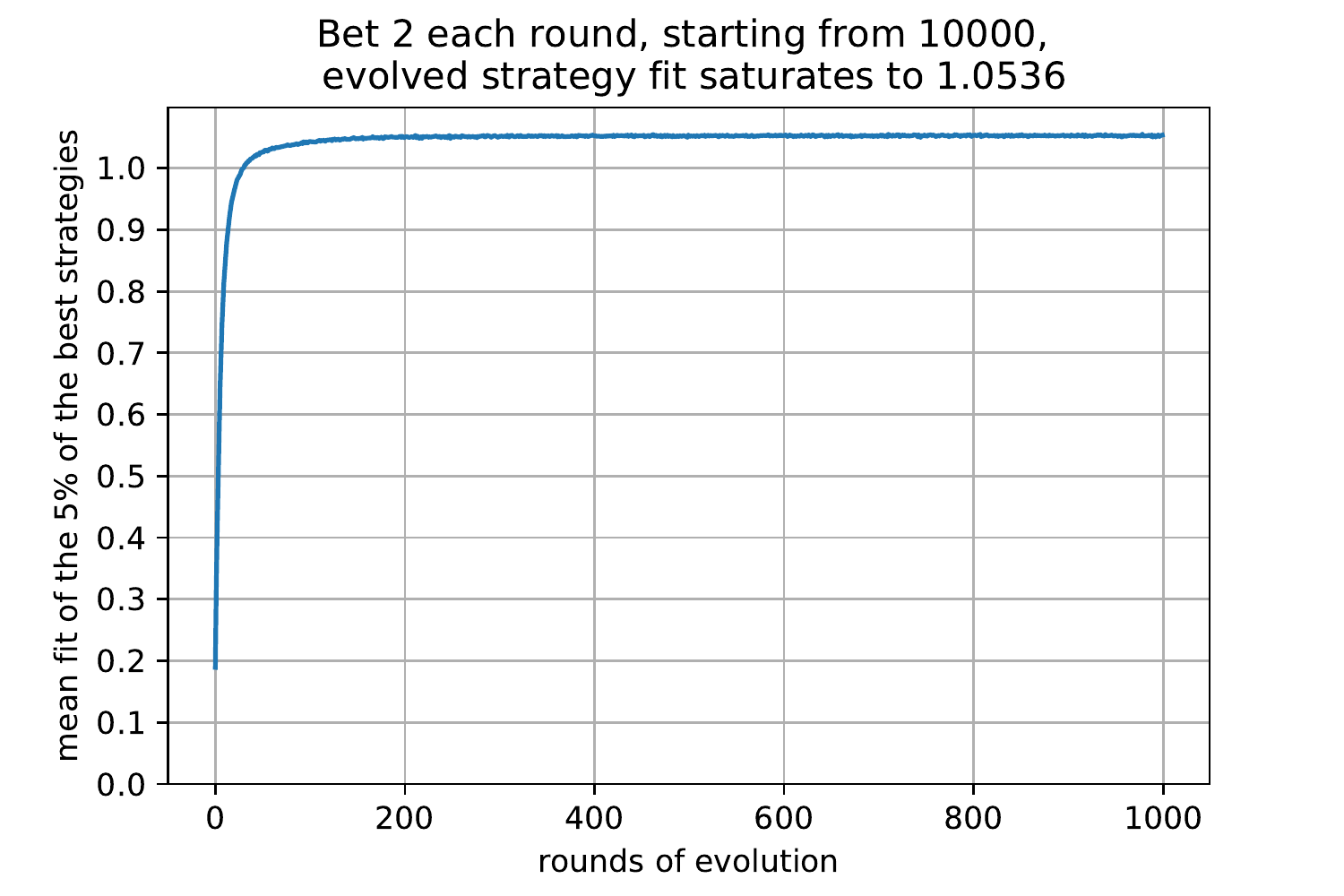}
 \caption{ \label{mean_fit_random} 
 Time dependence of the mean fit score, in the evolution of the random strategies in section \ref{section_evolving_random}.
 The fit score is defined as the return of the $\alpha=0.05$ of the most fit strategies, according to (\ref{mean_score_most_fit}).
 }
\end{center}
\end{figure}

\begin{figure}
\begin{center}
 \includegraphics[width=12cm,height=12cm,keepaspectratio]{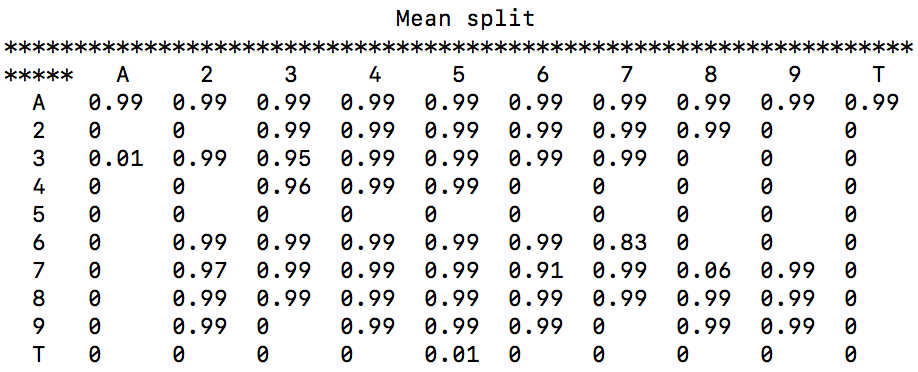}
 \caption{ \label{mean_split_random} 
 Mean split outcome in section \ref{section_evolving_random},
 starting from a population of 5000 random strategies after 1000 rounds of evolution.
 The columns represent the dealer's up-card, the rows represent the card in pair.
 The mean is taken over 5\% of the most fit strategies.
 }
 \end{center}
\end{figure}
\begin{figure}
\begin{center}
  \includegraphics[width=12cm,height=12cm,keepaspectratio]{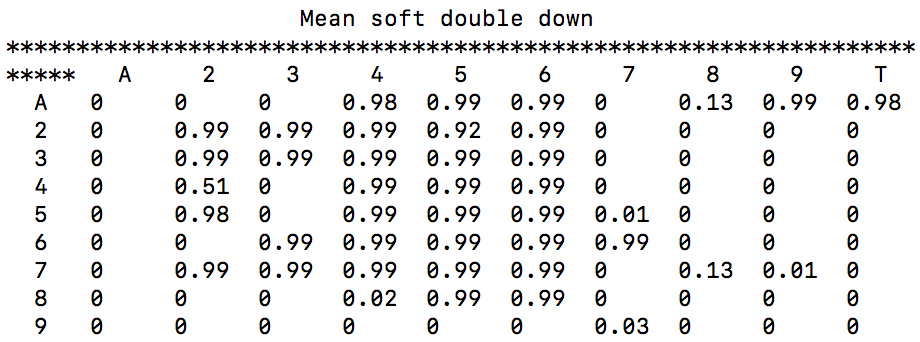}
 \caption{ \label{mean_soft_double_down} 
 Mean soft double down outcome in section \ref{section_evolving_random},
 starting from a population of 5000 random strategies after 1000 rounds of evolution.
 The columns represent the dealer's up-card, the rows represent the non-ace card of the player.
 The mean is taken over 5\% of the most fit strategies.
 }
 \end{center}
\end{figure}
 \begin{figure}
\begin{center}
  \includegraphics[width=12cm,height=12cm,keepaspectratio]{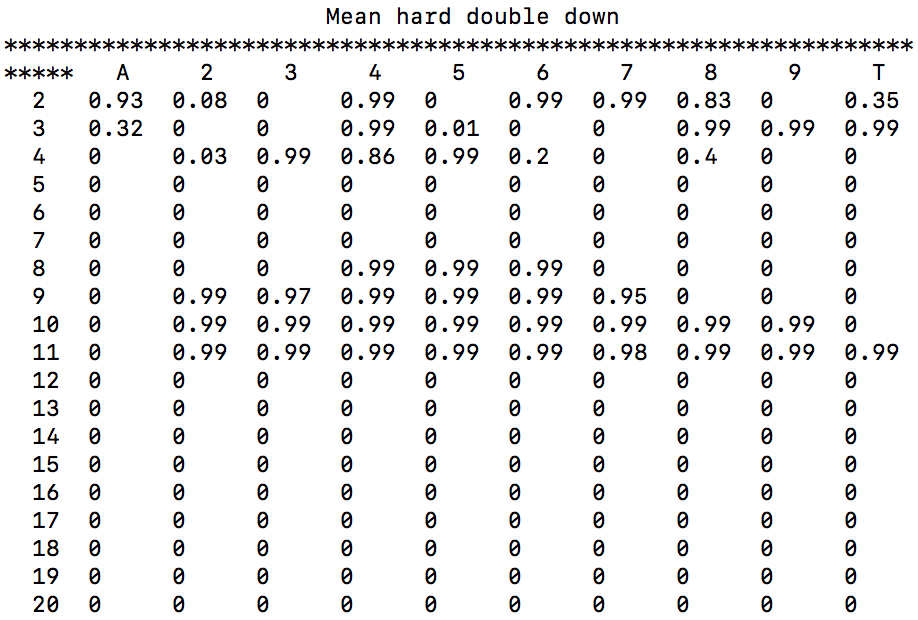}
 \caption{ \label{mean_hard_double_down} 
 Mean hard double down outcome in section \ref{section_evolving_random},
 starting from a population of 5000 random strategies after 1000 rounds of evolution.
 The columns represent the dealer's up-card, the rows represent the player's hand count.
 The mean is taken over 5\% of the most fit strategies.
 }
\end{center}
\end{figure}

\begin{figure}
\begin{center}
 \includegraphics[width=12cm,height=12cm,keepaspectratio]{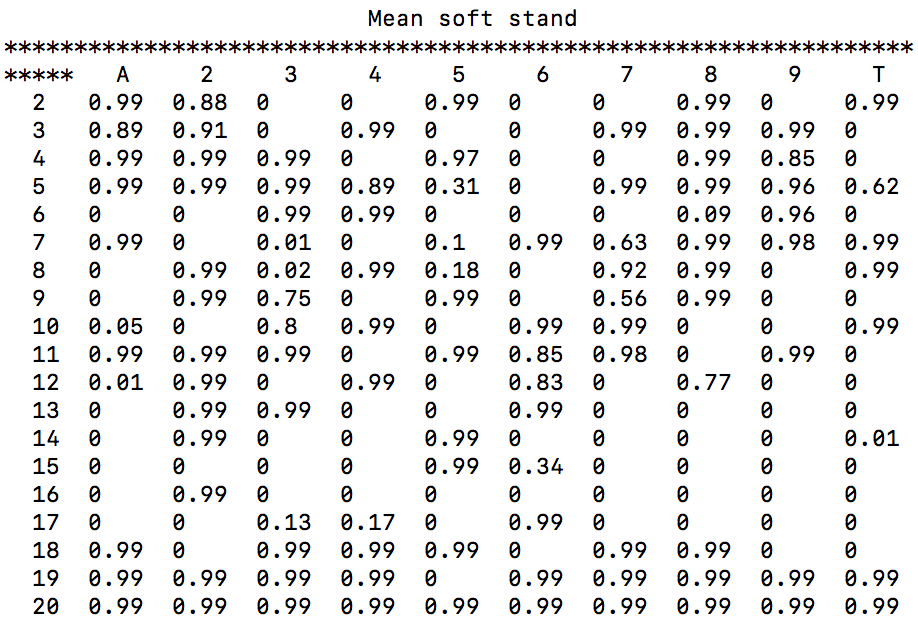}
 \caption{ \label{mean_soft_stand} 
 Mean soft stand outcome in section \ref{section_evolving_random},
 starting from a population of 5000 random strategies after 1000 rounds of evolution.
 The columns represent the dealer's up-card, the rows represent the player's hand count.
 The mean is taken over 5\% of the most fit strategies.
 }
 \end{center}
\end{figure}
 \begin{figure}
\begin{center}
  \includegraphics[width=12cm,height=12cm,keepaspectratio]{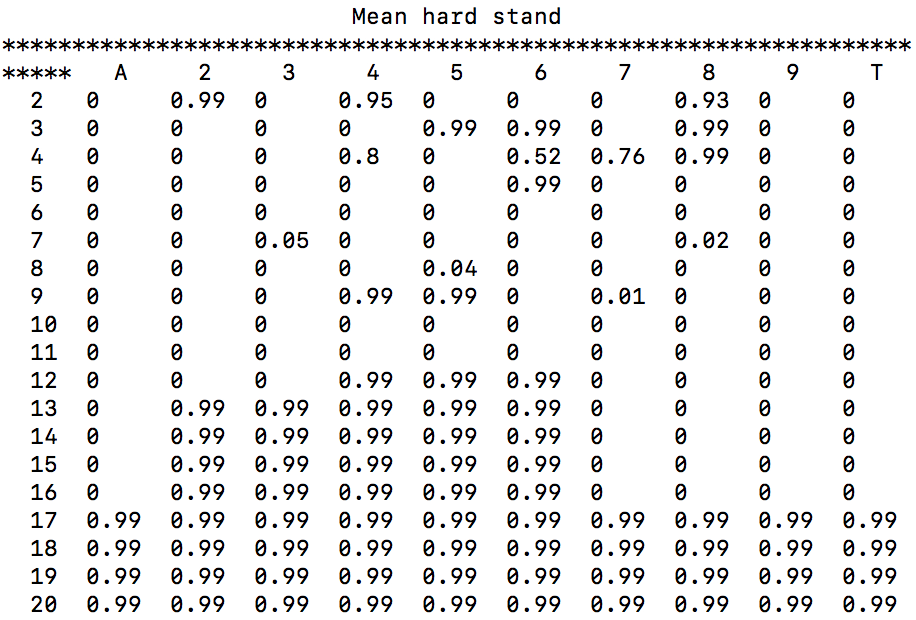}
 \caption{ \label{mean_hard_stand} 
 Mean hard stand outcome in section \ref{section_evolving_random},
 starting from a population of 5000 random strategies after 1000 rounds of evolution.
 The columns represent the dealer's up-card, the rows represent the player's hand count.
 The mean is taken over 5\% of the most fit strategies.
 }
\end{center}
\end{figure}

In this section we study evolution of the population of blackjack strategies which have
been initialized randomly. The evolution is done according to the algorithm
discussed in section \ref{evolutionary_discussion_section}. As we will see in this section,
the impressive result is that the evolution relatively quickly finds a strategy which performs
(almost) as well as the Thorp's basic strategy, discussed in section \ref{thorp_section},
and agrees quite with the Thorp's basic strategy in some of the specifics.

To make an exhaustive and complete search we run the evolution over $\tau=1000$ generations.
We use the evolutionary selection parameters as described in section \ref{evolutionary_discussion_section},
that is, we use the selection rate (\ref{alpha_rate}), and the propagation rate (\ref{propagation_rate}).
We evolve the population of $M=5000$ strategies, and reset the bankroll of all the strategies to $\$ 10^4$
at the beginning of each generation.
\footnote{
If we don't reset the bankroll at the beginning of each evolutionary round, the bankroll score of the most fit strategies
will go into the linear regime after the large number of evolutionary steps.
Also, in the beginning of evolution, the sub-optimal but lucky strategies will get a persisting undeserved advantage 
in the selection, creating a noise and slowing down the evolution.} The resulting mean fit score (\ref{mean_score_most_fit})
is plotted in figure \ref{mean_fit_random}. 

Notice that the saturation value of the fit score in figure \ref{mean_fit_random}
indicates that the fit strategies make about $\$ 540$
after $T\simeq 10^4$ rounds of betting $\$ 2$. The apparent discrepancy from the more modest result of making
about $\$ 50$ by playing the Thorp's
basic strategy (see figure \ref{thorp_player_bankroll}) is due to the artifact of using the highest performed $5\%$ 
of the 5000 strategies in the given round of evolution to calculate the mean return. From the distribution of the player's finishes in figure \ref{thorp_player_bankroll}
one can see that the top $5\%$ indeed make about $\$ 500$ on average.

We output the mean evolved basic strategy by calculating the weighted average of the top $\alpha M=250$ strategies.
For the weights we use the fit scores of those strategies. The results for the split, soft/hard double down,
and the soft/hard stand evolved mean strategies are provided in figures \ref{mean_split_random}, \ref{mean_soft_double_down},
\ref{mean_hard_double_down}, \ref{mean_soft_stand}, \ref{mean_hard_stand}.

As was pointed out in section \ref{evolutionary_discussion_section},
some of the genes in the selected strategy parametrization have no phenotypic expression.
For instance, the first ten rows in the mean soft stand table \ref{mean_soft_stand} have no influence
on the way the game which uses that table is played. This is because the soft hand by definition contains
an ace which can be counted as 11, and therefore its score can be only 12 and higher. It is illustrative to observe
how some of those genes on average have the values of 0.99/0, seemingly asserting to a significance of having 
1/0 in that location. This is, however, an example of how a gene can survive just because it happens to be good
at replication: a chance occurrence of having a majority of 1/0 in that chromosome location in the given subset of the most fit strategies
might reinforce itself and converge to having a uniformly found 1/0 in that chromosome slot. This statement is confirmed by
running repeated evolution simulations and observing that the values of 1/0 are found randomly in
different chromosome slots (in the subset of genes with no phenotypic expression), for different evolutionary outcomes.

\begin{figure}
\begin{center}
\includegraphics[width=6cm, height=3.75cm]{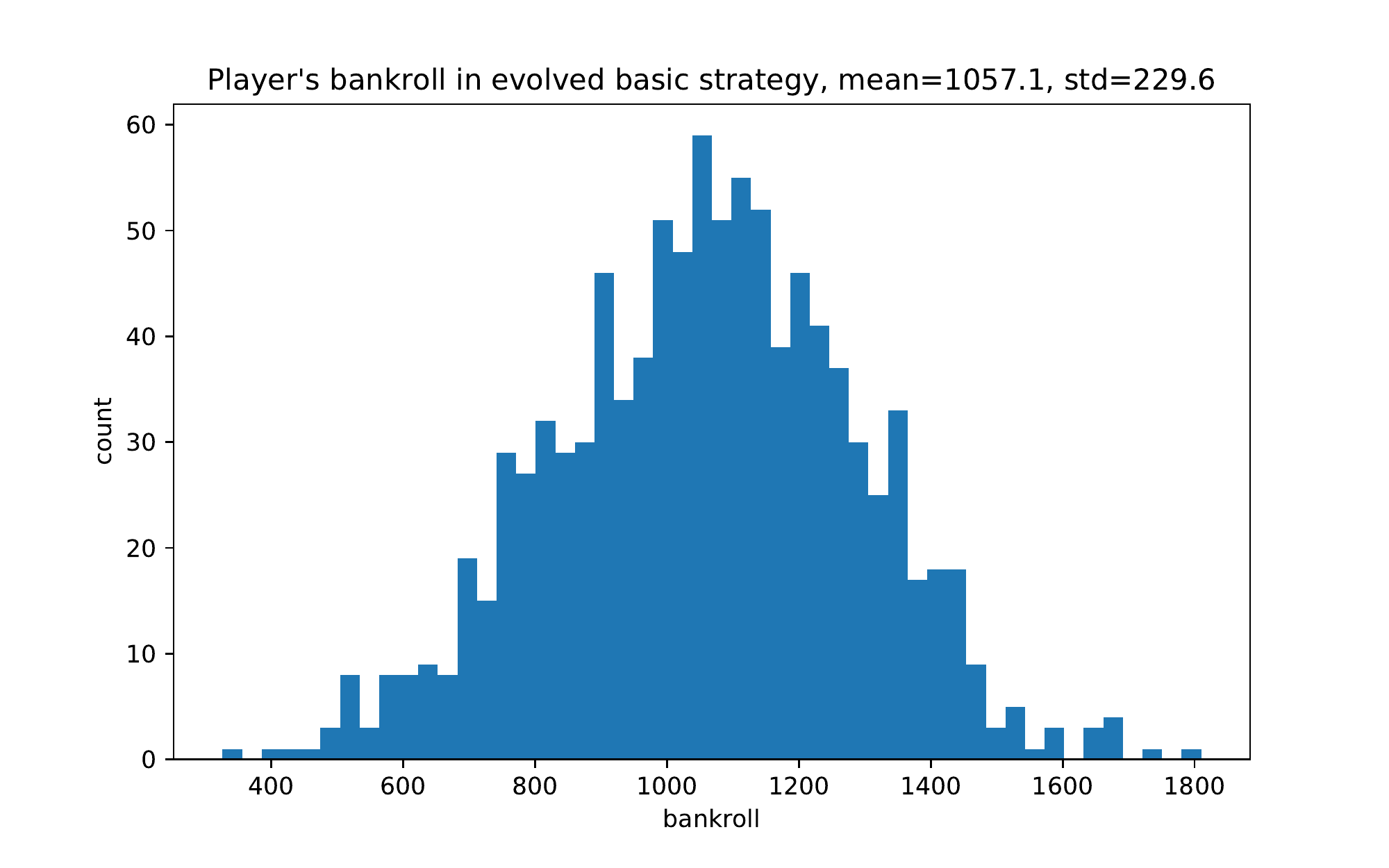}
 \includegraphics[width=6cm, height=3.75cm]{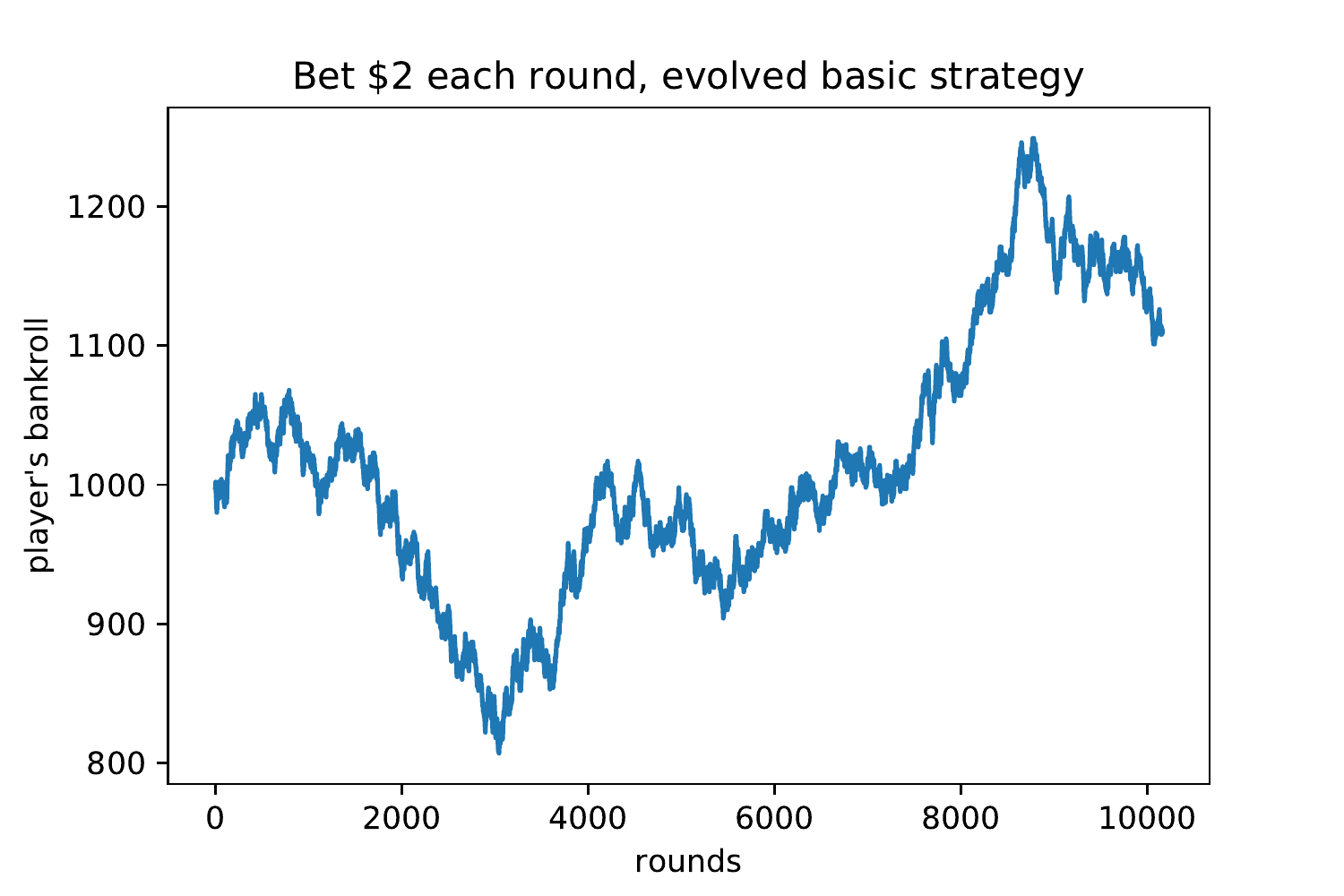}
 \caption{ \label{evolved_player_bankroll} 
 {\bf Left:} Player's final bankroll after 10000 rounds of game (with about 200 extra split rounds),
 {\bf Right:} Player's bankroll time series in one of the games; played according to evolved basic strategy
 (see figures  \ref{mean_split_random}, \ref{mean_soft_double_down},
\ref{mean_hard_double_down}, \ref{mean_soft_stand}, \ref{mean_hard_stand}, with entries taken to be $1$
for the mean larger than or equal to $0.95$) in the simulation described in section~\ref{section_evolving_random}.
 Player starts with \$1000 bankroll ands bets \$2 per round.
 }
\end{center}
\end{figure}

\begin{figure}
\begin{center}
\includegraphics[width=6cm, height=3.75cm]{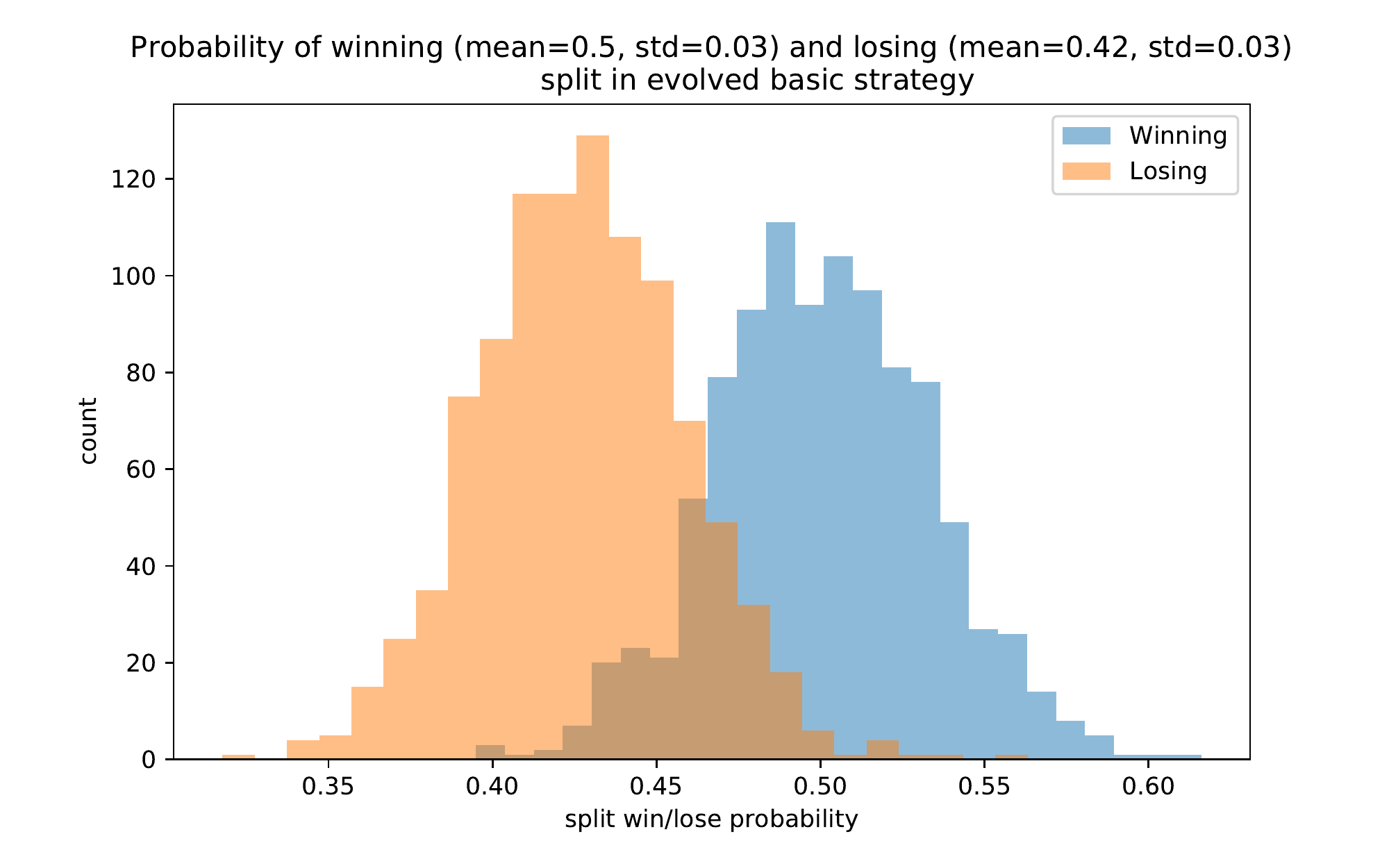}
 \includegraphics[width=6cm, height=3.75cm]{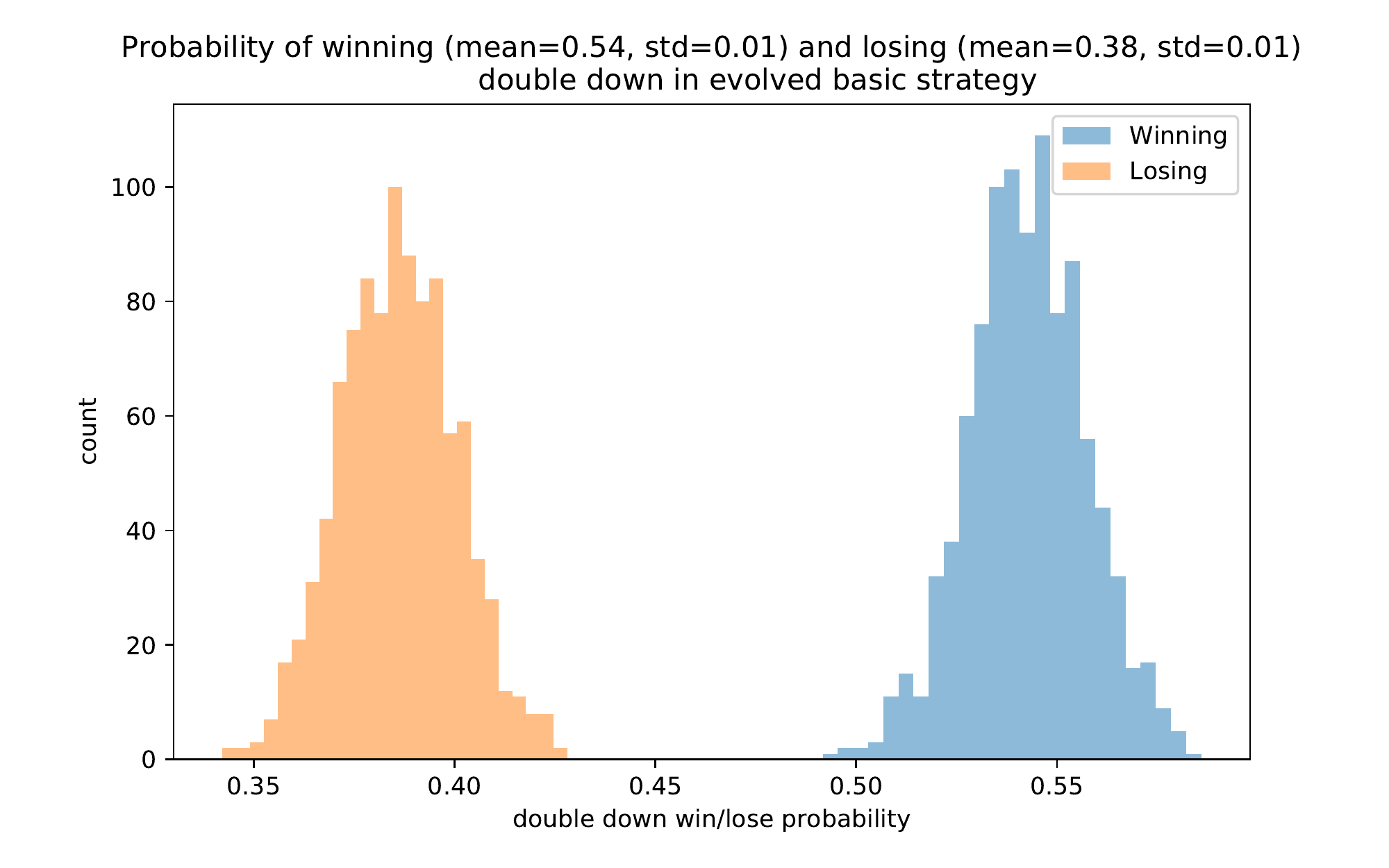}
 \caption{ \label{evolved_split_dd_win_loss} 
 Probability of winning/losing split ({\bf left}) and double down ({\bf right}) in evolved basic strategy,
 see figures  \ref{mean_split_random}, \ref{mean_soft_double_down},
\ref{mean_hard_double_down}, \ref{mean_soft_stand}, \ref{mean_hard_stand}, with entries taken to be $1$
for the mean larger than or equal to $0.95$) in the simulation described in section~\ref{section_evolving_random}.
 The mean probability of split is 0.01, the mean probability of double down is 0.12.
 }
\end{center}
\end{figure}



Taking into account irrelevance of the spurious genes with no phenotypic expression we then
notice that the standing prescription for both the soft and the hard hands agree well
for the Thorp's strategy, see table \ref{thorp_stand}, and the evolved strategy, see figures \ref{mean_soft_stand}, \ref{mean_hard_stand}.
Comparing these strategy tables we should also keep in mind that the decisions to split and to double down, when
splitting and doubling down is incorporated into the
basic strategy, is to be done before the decision to hit/stand.
For instance, the evolved strategy prescribes to double down on (A,7) against dealer's 2, on (A,8) against 
dealer's 5, and on (A,7) against dealer's 6, see figure \ref{mean_soft_double_down}.
This will suppress significance of expression of zero entries for the player's soft count 18 against dealer's 2,
19 against dealer's 5, and 18 against dealer's 6,
in the soft stand table \ref{mean_soft_stand}. One might wonder what happens
when, for instance, we consider standing with soft 18 made of three cards,
such as (A,5,2), against dealer's 6, in which case the decision to double
down is not considered (player can double down only on the originally
dealt hand of two cards). In such cases we notice, however, that the player will
in fact likely double down {\it before} getting to that hand, say, both (A,5) and (A,2) against dealer's 6
are prescribed to double down on by the table \ref{mean_soft_double_down}.

Similarly the apparent standing prescriptions,
which the soft stand table \ref{mean_soft_stand} indicates for player's
count against dealer's up-card, at (13,2), (14,2), (16,2), (13,3), (14,5), (15,5), (13,6), (17,6),
are suppressed due to the corresponding priority of the soft double down prescriptions
by the table \ref{mean_soft_double_down}. At the same time, splitting prescription for
aces, due to figure \ref{mean_split_random}, suppresses significance of the soft stand
entries (12,2) and (12,4) of the table \ref{mean_soft_stand}.
We also notice that the (9,4) and (9,5) prescriptions of the hard stand table \ref{mean_hard_stand}
are suppressed by the prescriptions to double down on hard 9 against dealer's 4 and 5, due to the
hard double down table \ref{mean_hard_double_down}. The hard stand prescription at (4,8) in
figure \ref{mean_hard_stand} will be suppressed by the prescription to split pair of 2 against dealer's 8 due to
figure \ref{mean_split_random}.

We also ran a separate calculation for 100 evolutionary steps, for the strategies which
never split and never double down.
The rest of the evolutionary conditions were the same as set above.
This allows to test whether the evolutionary selection can find the
correct standing prescription, and avoid having standing prescriptions being suppressed
by the split and double down prescriptions, as discussed above. We confirmed that the resulting evolved strategy has the standing prescription
agreeing with the Thorp's prescription, shown in table \ref{thorp_stand}. This suggest that
an alternative way to learn the basic strategy evolutionary would be to evolve
the correct standing, doubling down, and splitting prescriptions separately, in that order.
For instance, after having learned the correct standing prescription, we could use it while
learning the correct prescription for doubling down (while still omitting the splits).
We will not
be using such an approach in this paper, preferring to evolve the whole strategy simultaneously.

We then notice that while some of the prescriptions for the split and the soft/hard double down agree for the Thorp's
and the evolved basic strategies (for instance, the evolved basic strategy has learned
to always split aces and never split fives and tens), some distinctions nevertheless exist.\footnote{
For instance, in agreement with \cite{Fogel2004} our strategies learned to doubled down on the (A,8)
against dealer's 5 and 6,
as contrasted with the Thorp's basic strategy.
}
Some of the apparent distinctions are spurious, for instance, the apparent prescription
to double down on hard 4 against dealer's 3 and 5, indicated by figure \ref{mean_hard_double_down},
is suppressed by prescription of figure \ref{mean_split_random} to split pair of 2,
against dealer's 3 and 5.

To analyze an impact of more essential distinctions
we run simulation tests of the evolved strategy, similar to those we did in section \ref{thorp_section}
for the Thorp's basic strategy. First we need to decide on how to choose the basic strategy from the mean
strategies \ref{mean_split_random}, \ref{mean_soft_double_down},
\ref{mean_hard_double_down}, \ref{mean_soft_stand}, \ref{mean_hard_stand}. The most straightforward way is to
take the entry value to be 1 if the corresponding mean gene is greater than or equal to some threshold value $a$.
In this section we choose $a=0.95$ for such a threshold.

We present the results in figures \ref{evolved_player_bankroll}, \ref{evolved_split_dd_win_loss}.
Comparing figures \ref{thorp_split_dd_win_loss} and \ref{evolved_split_dd_win_loss}
for the Thorp's basic strategy and the evolved basic strategy split performance, we notice
that the split prescription in the evolved strategy has a higher winning edge ($8\%$)
than its counterpart in the Thorp's strategy ($5\%$). However, the evolved strategy split prescription
results in splitting of about 1\% of the hands, while the Thorp's split strategy results in
splitting of about 2\% of the hands.

From figure \ref{evolved_player_bankroll} we see that the mean player's
bankroll, when using the evolved basic strategy, is comparable to the Thorp's result.
Specifically, the player's final bankroll mean and standard deviation are $B_T=\$ 1057.1$,
$s=\$ 229.6$, giving the $95\%$ confidence interval for the final bankroll $[\$ 1042.9, \$ 1071.3]$,
and for the player's edge, $[0.19\%,0.32\%]$.

\begin{figure}
\begin{center}
\includegraphics[width=12cm, height=7.5cm]{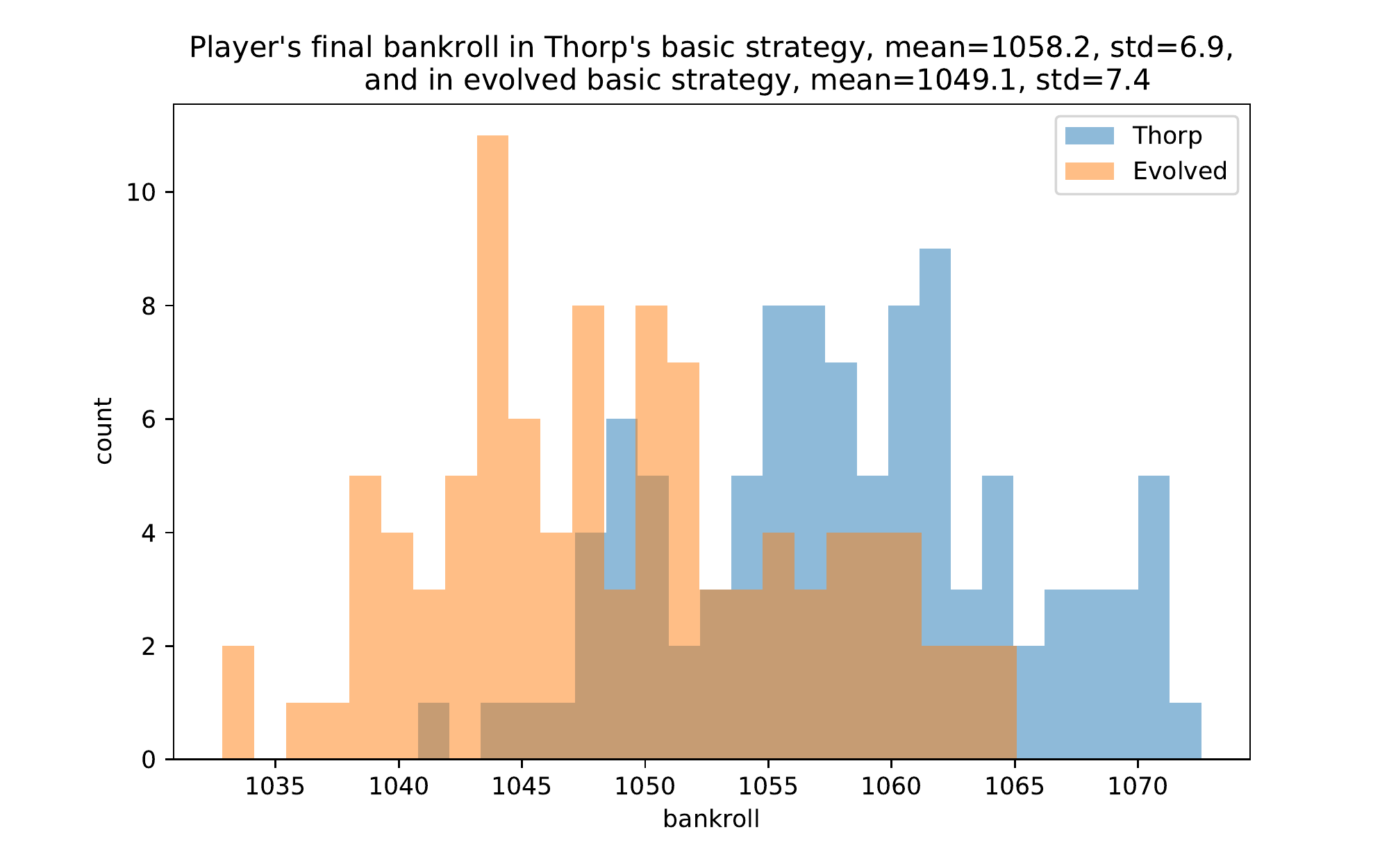}
 \caption{ \label{evolved_thorp_performance} 
 Comparison of the Thorp's basic strategy and the evolved basic strategy.
 Both histograms are obtained by running 100 simulations
 where in each simulation we play 1000 games and calculate the mean
 final bankroll (over the entire population).
 }
\end{center}
\end{figure}

We can perform a $t$-test quantifying the difference in performance of the Thorp's
basic strategy and the evolved basic strategy.
We perform the test of the hypothesis that the two distributions have the same mean.
To this end we calculate the
$t$-statistic
\begin{equation}
\label{t_mean_difference}
t=\frac{\bar{B}_{{\rm Thorp}}-\bar{B}_{{\rm Evolved}}}{s\,\sqrt{\frac{1}{n_{{\rm Thorp}}
}+\frac{1}{n_{{\rm Evolved}}}}}=-0.55\,,
\end{equation}
where we substituted the pooled standard deviation (assuming that both distributions
have the same standard deviation)
\begin{equation}
s=\sqrt{\frac{(n_{{\rm Thorp}}-1)s_{{\rm Thorp}}^2+(n_{{\rm Evolved}}-1)s_{{\rm Evolved}}^2}
{n_{{\rm Thorp}}+n_{{\rm Evolved}}-2}}\,,
\end{equation}
with the parameters
\begin{align}
\bar{B}_{{\rm Thorp}}&=\$1051.4\,,\quad s_{{\rm Thorp}}=\$232.0\,,\quad n_{{\rm Thorp}}=1000\,,\\
\bar{B}_{{\rm Evolved}}&=\$1057.1\,,\quad s_{{\rm Evolved}}=\$229.6\,,\quad n_{{\rm Evolved}}=1000\,,
\end{align}
The $t$ in (\ref{t_mean_difference}) has $t$-distribution with $n_{{\rm Thorp}}+ n_{{\rm Evolved}}-2=1998$
degrees of freedom, giving about $42\%$ confidence that the mean values $\bar{B}_{{\rm Thorp}}$
and $\bar{B}_{{\rm Evolved}}$ come from distributions with different means. This is in no way
an assertive statement that the evolved strategy has a different performance mean than the
Thorp's basic strategy.

Running 100 simulations, equivalent to those used to produce figure \ref{evolved_player_bankroll},
and averaging out the so-obtained 100 means,
we obtain the mean bankroll $\$ 1049.1$, and the standard deviation $\$ 7.4$.
This is to be compared with the Thorp's performance, calculated in section \ref{thorp_section},
which gives the mean $\$ 1058.2$, and the standard
deviation of $\$ 6.9$, as we illustrate
in figure \ref{evolved_thorp_performance}.

Our results for both the Thorp's and the evolved strategy similarly successful 
performance
suggest that the there is an entire region in the parameter space of the basic strategy,
containing the strategies performing similarly well. This is true at least
when these strategies are considered in the framework with
the given bet size, $b=\$2$, and the number of rounds played, $N=10^4$, used to quantify the performance.
When the bet size $b$ and the number of rounds played $N$ is taken to be larger, penalization of the
deviation from the most optimal basic strategy can be made as large as possible.
In other words, the gradient of the basic strategy fit function $\phi$, see (\ref{fit_rank}), in the direction
of any given parameter of the basic strategy
should be sufficiently large for the evolutionary pressure to act on that parameter.

We confirm this by testing both the Thorp's basic strategy and the evolved
basic strategy by running simulations with $N=10^5$ rounds of game, with $b=\$2$ bet per game.
The player starts with $\$5000$. The mean final bankroll over $M=1000$ games
is $\$5578.3$, and the standard deviation is $\$ 724.9$ for the Thorp's basic strategy.
For the evolved basic strategy the mean final bankroll is $\$ 5455.4$,
and the standard deviation is $\$ 705.1$. The corresponding $t$-value
of the statistic (\ref{t_mean_difference}) is $3.8$, giving $99.9\%$ confidence
that these two bankroll values come from distributions with different means,
agreeing with what is shown in figure \ref{evolved_thorp_performance}.

\section{Evolving Thorp's basic strategy}
\label{section_evolving_thorp}

\begin{figure}
\begin{center}
 \includegraphics[width=12cm,height=12cm,keepaspectratio]{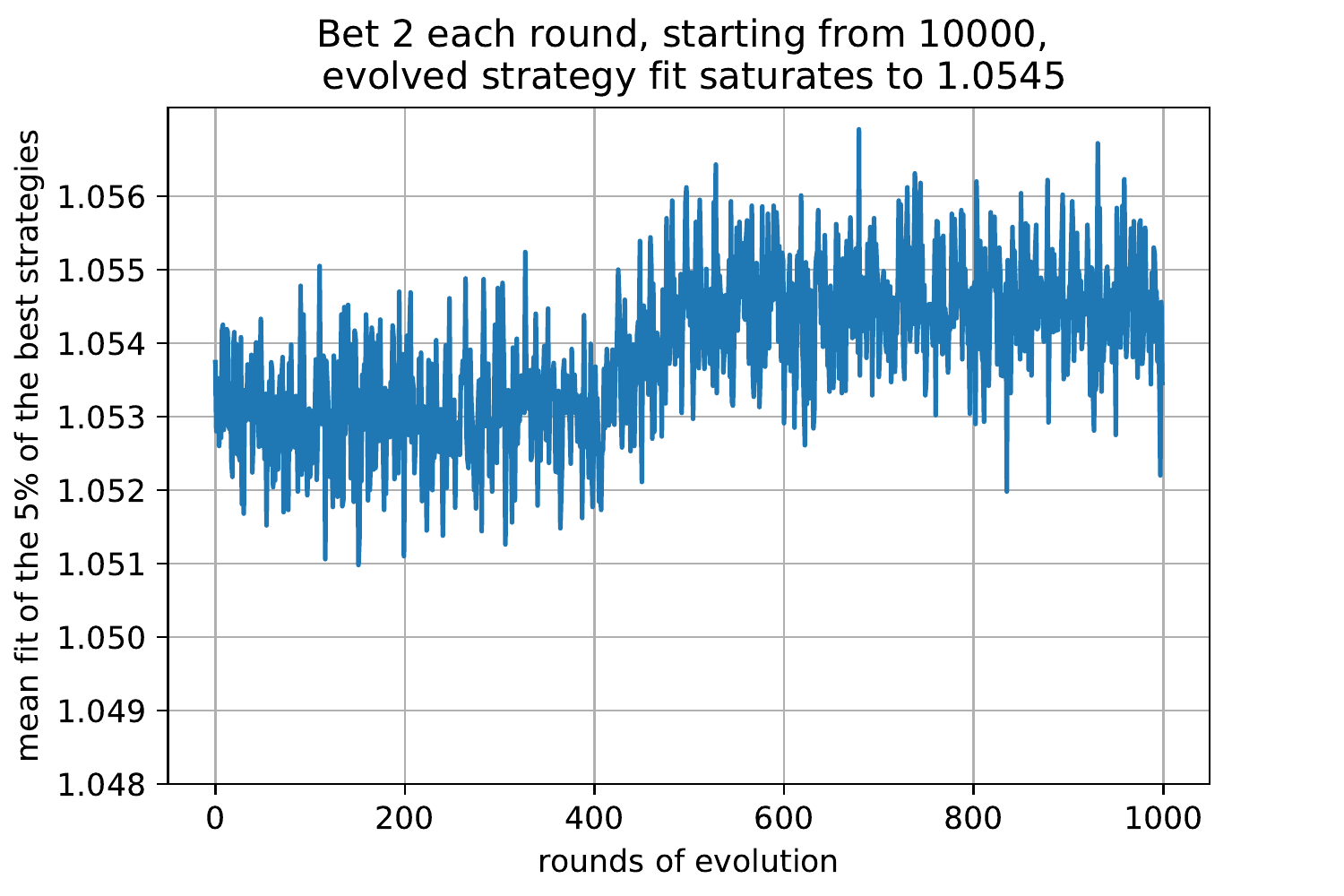}
 \caption{ \label{mean_fit_random_thorp} 
 Time dependence of the mean fit score, in the evolution of the Thorp's strategies in section \ref{section_evolving_thorp}.
 The fit score is defined as the return of the $\alpha=0.05$ of the most fit strategies, according to (\ref{mean_score_most_fit}).
 }
\end{center}
\end{figure}

\begin{figure}
\begin{center}
 \includegraphics[width=12cm,height=12cm,keepaspectratio]{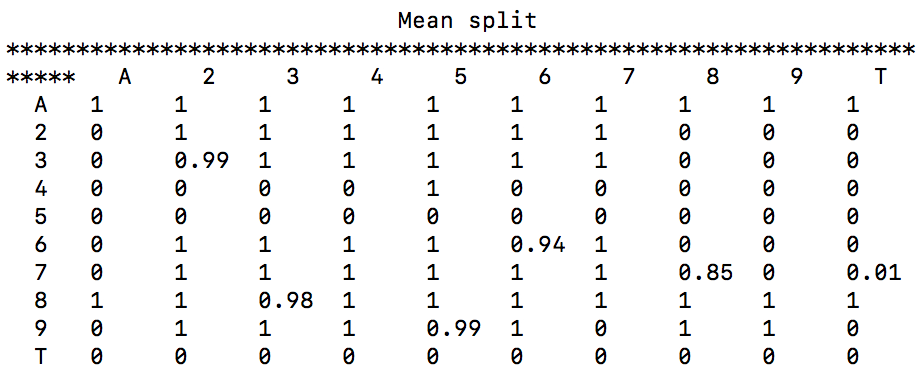}
 \caption{ \label{mean_split_evolved_thorp} 
 Mean split outcome in section \ref{section_evolving_thorp},
 starting from a population of 5000 Thorp's strategies after 1000 rounds of evolution.
 The columns represent the dealer's up-card, the rows represent the card in pair.
 The mean is taken over 5\% of the most fit strategies.
 }
 \end{center}
\end{figure}
 \begin{figure}
\begin{center}
  \includegraphics[width=12cm,height=12cm,keepaspectratio]{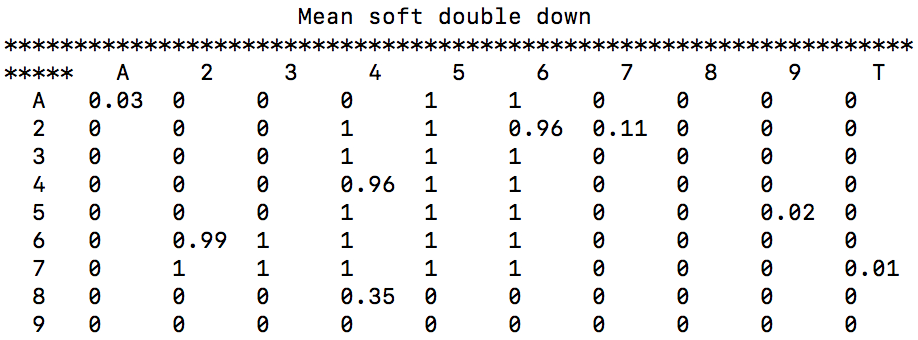}
 \caption{ \label{mean_soft_double_down_evolved_thorp} 
 Mean soft double down outcome in section \ref{section_evolving_thorp},
 starting from a population of 5000 Thorp's strategies after 1000 rounds of evolution.
 The columns represent the dealer's up-card, the rows represent the non-ace card of the player.
 The mean is taken over 5\% of the most fit strategies.
 }
 \end{center}
\end{figure}
 \begin{figure}
\begin{center}
  \includegraphics[width=12cm,height=12cm,keepaspectratio]{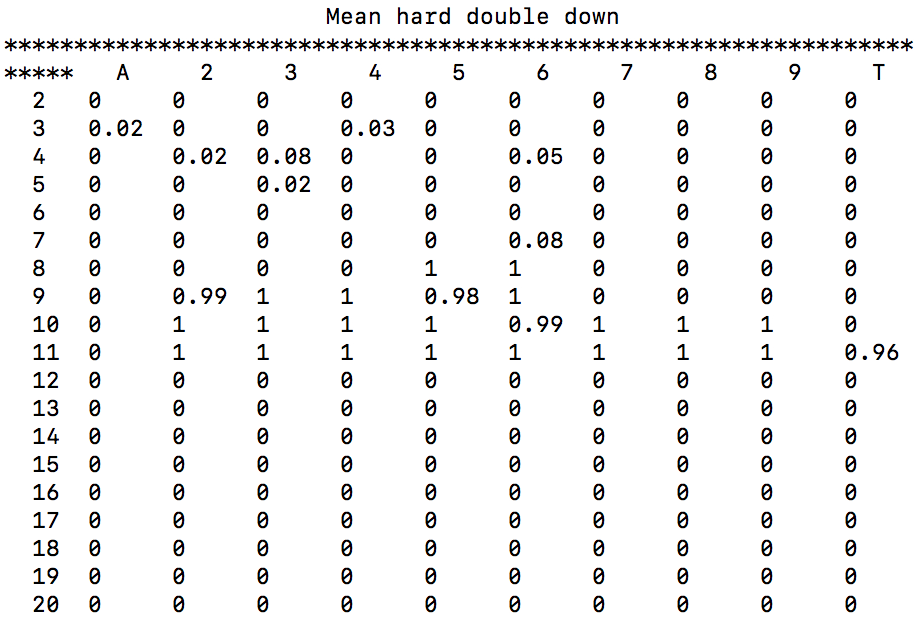}
 \caption{ \label{mean_hard_double_down_evolved_thorp} 
 Mean hard double down outcome in section \ref{section_evolving_thorp},
 starting from a population of 5000 Thorp's strategies after 1000 rounds of evolution.
 The columns represent the dealer's up-card, the rows represent the player's hand count.
 The mean is taken over 5\% of the most fit strategies.
 }
\end{center}
\end{figure}

\begin{figure}
\begin{center}
 \includegraphics[width=12cm,height=12cm,keepaspectratio]{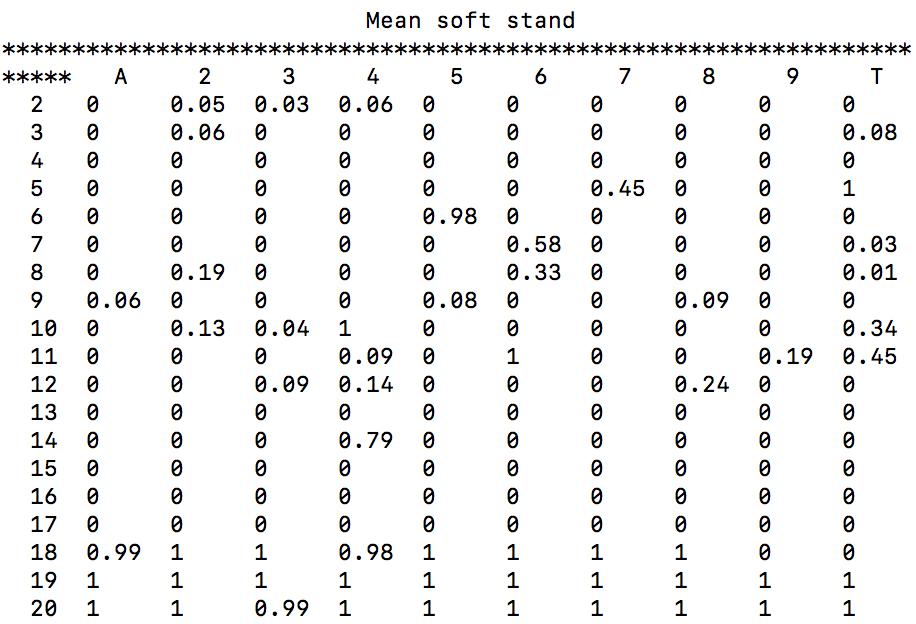}
 \caption{ \label{mean_soft_stand_evolved_thorp} 
 Mean soft stand outcome in section \ref{section_evolving_thorp},
 starting from a population of 5000 Thorp's strategies after 1000 rounds of evolution.
 The columns represent the dealer's up-card, the rows represent the player's hand count.
 The mean is taken over 5\% of the most fit strategies.
 }
 \end{center}
\end{figure}
\begin{figure}
\begin{center}
  \includegraphics[width=12cm,height=12cm,keepaspectratio]{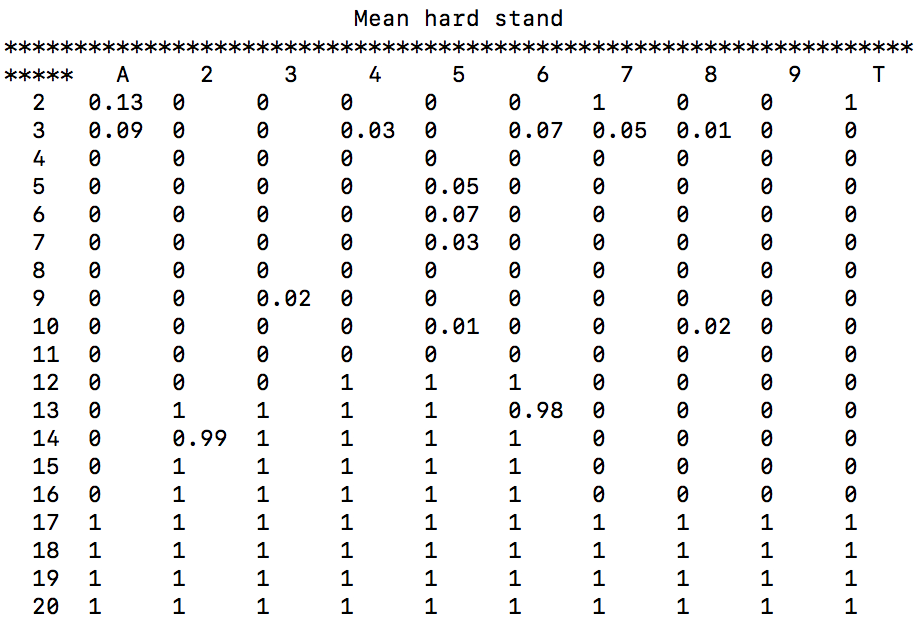}
 \caption{ \label{mean_hard_stand_evolved_thorp} 
 Mean hard stand outcome in section \ref{section_evolving_thorp},
 starting from a population of 5000 Thorp's strategies after 1000 rounds of evolution.
 The columns represent the dealer's up-card, the rows represent the player's hand count.
 The mean is taken over 5\% of the most fit strategies.
 }
\end{center}
\end{figure}

\begin{figure}
\begin{center}
\includegraphics[width=6cm, height=3.75cm]{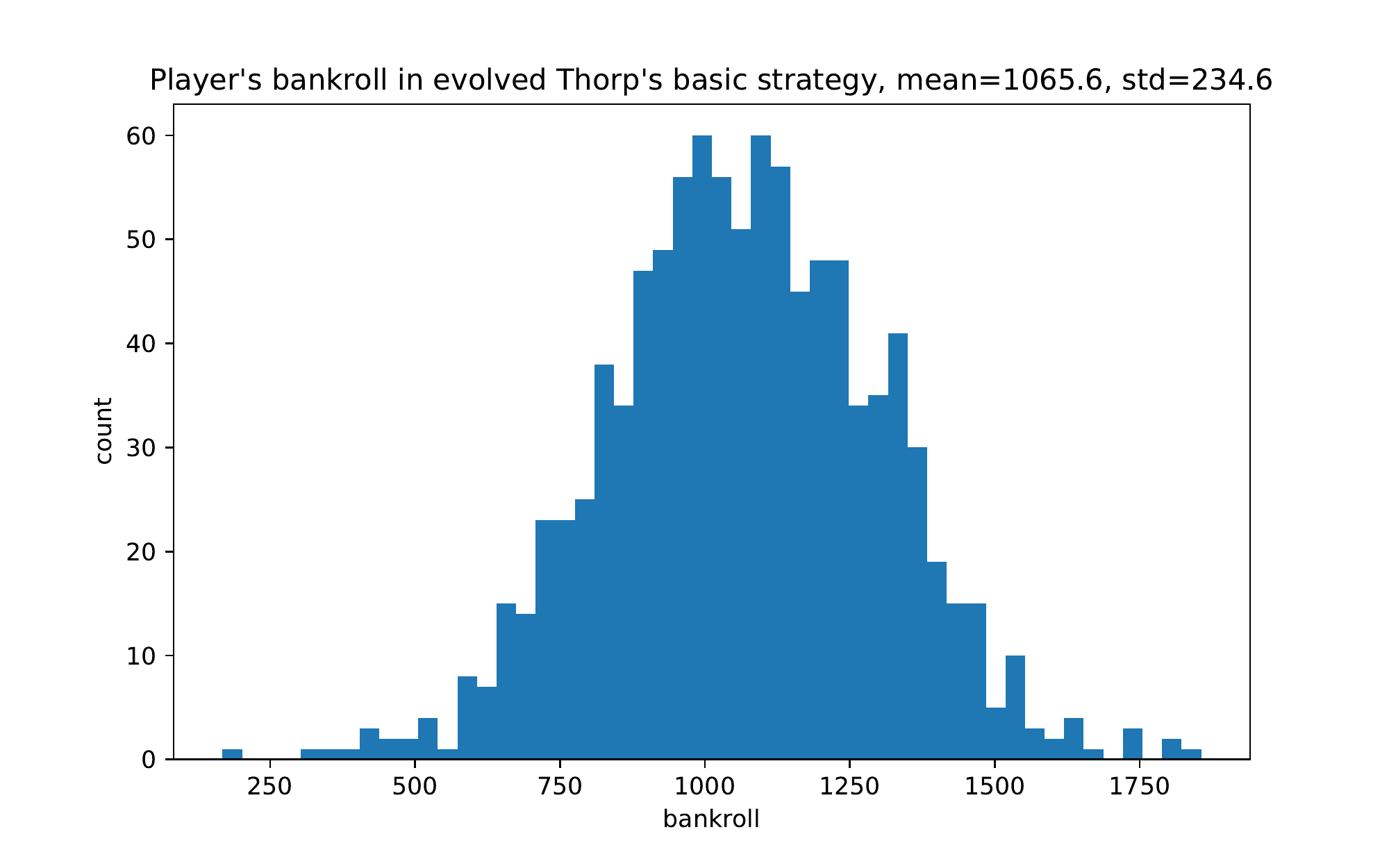}
 \includegraphics[width=6cm, height=3.75cm]{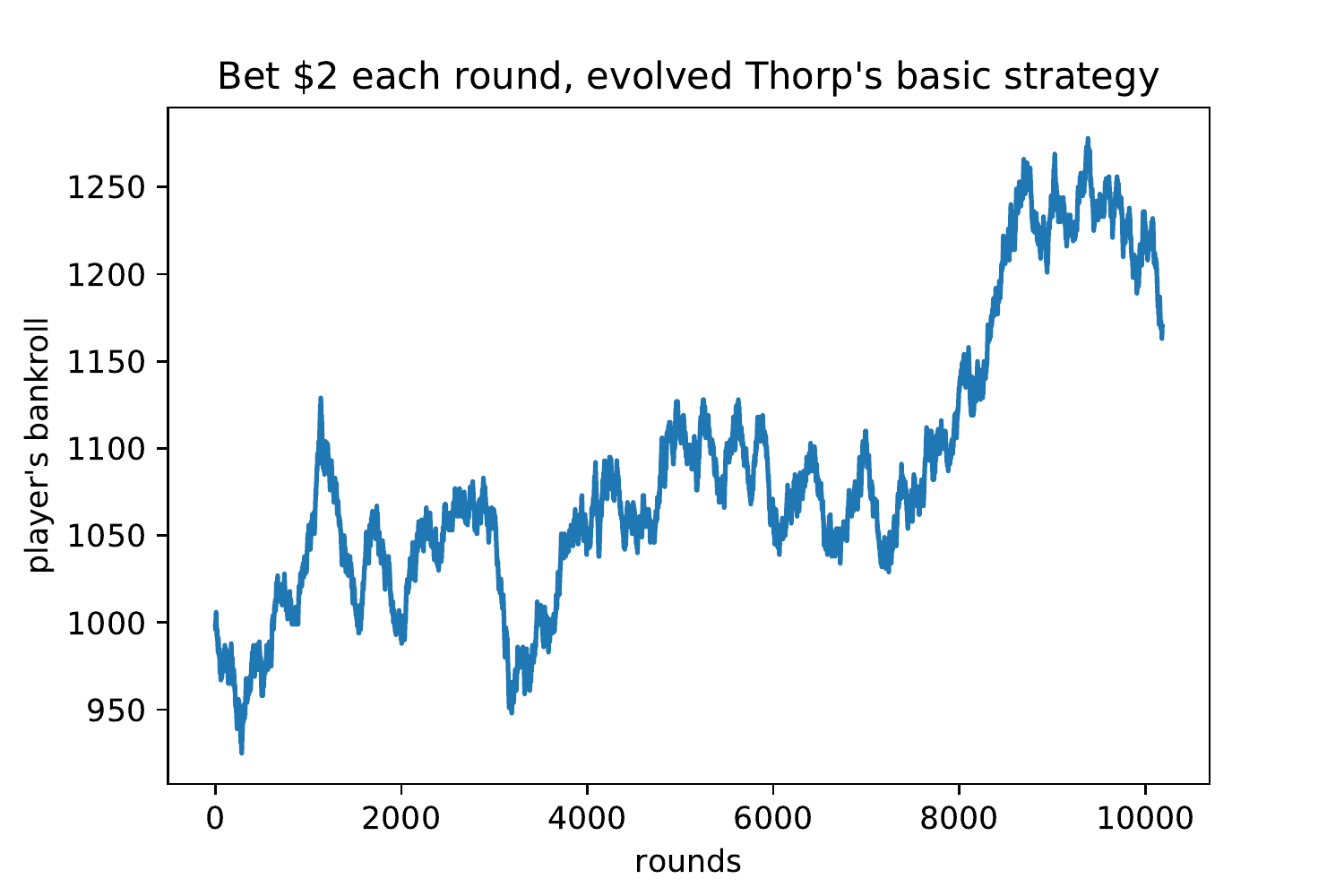}
 \caption{ \label{evolved_thorp_player_bankroll} 
 {\bf Left:} Player's final bankroll after 10000 rounds of game (with about 200 extra split rounds),
 {\bf Right:} Player's bankroll time series in one of the games; played according to evolved Thorp's basic strategy
 (see figures  \ref{mean_split_evolved_thorp}, \ref{mean_soft_double_down_evolved_thorp},
\ref{mean_hard_double_down_evolved_thorp}, \ref{mean_soft_stand_evolved_thorp}, \ref{mean_hard_stand_evolved_thorp}, with entries taken to be $1$
for the mean larger than or equal to $0.95$) in the simulation described in section~\ref{section_evolving_thorp}.
 Player starts with \$1000 bankroll ands bets \$2 per round.
 }
\end{center}
\end{figure}

\begin{figure}
\begin{center}
 \includegraphics[width=6cm, height=3.75cm]{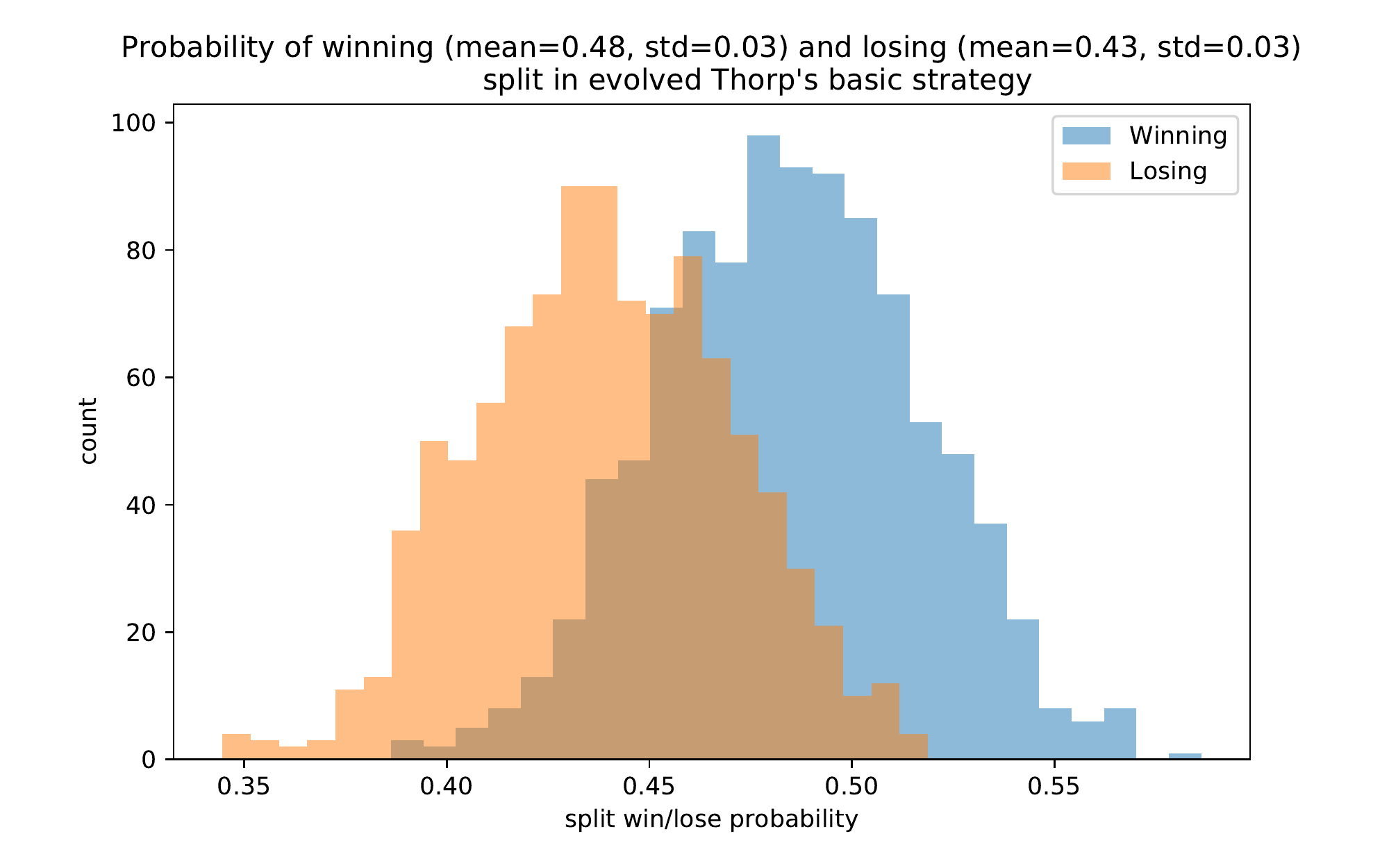}
\includegraphics[width=6cm, height=3.75cm]{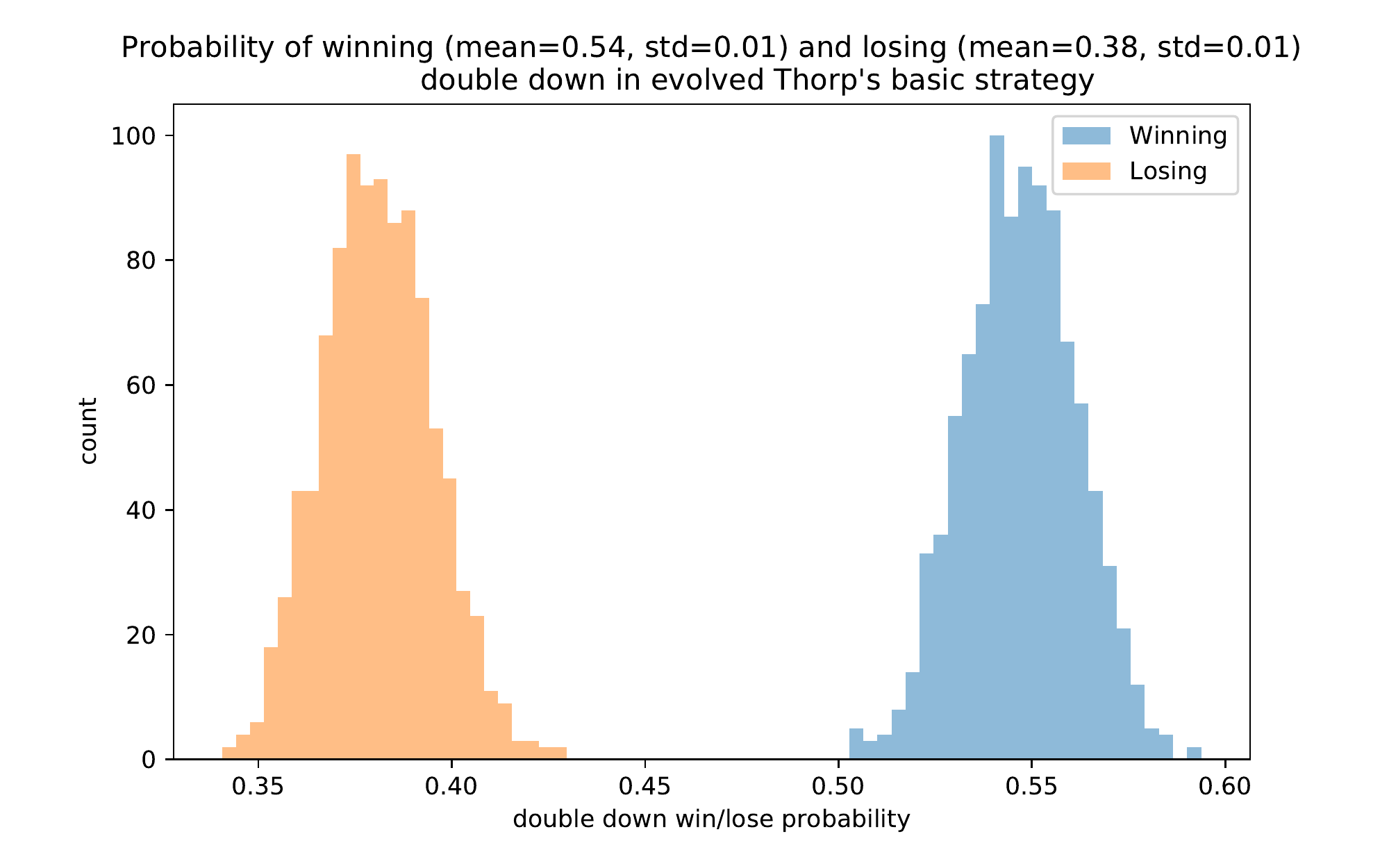}
 \caption{ \label{evolved_thorp_split_dd_win_loss} 
 Probability of winning/losing split ({\bf left}) and double down ({\bf right}) in evolved Thorp's basic strategy,
 see figures  \ref{mean_split_evolved_thorp}, \ref{mean_soft_double_down_evolved_thorp},
\ref{mean_hard_double_down_evolved_thorp}, \ref{mean_soft_stand_evolved_thorp}, \ref{mean_hard_stand_evolved_thorp}, with entries taken to be $1$
for the mean larger than or equal to $0.95$) in the simulation described in section~\ref{section_evolving_thorp}.
 The mean probability of split is 0.01, the mean probability of double down is 0.11.
 }
\end{center}
\end{figure}

\begin{figure}
\begin{center}
\includegraphics[width=12cm, height=7.5cm]{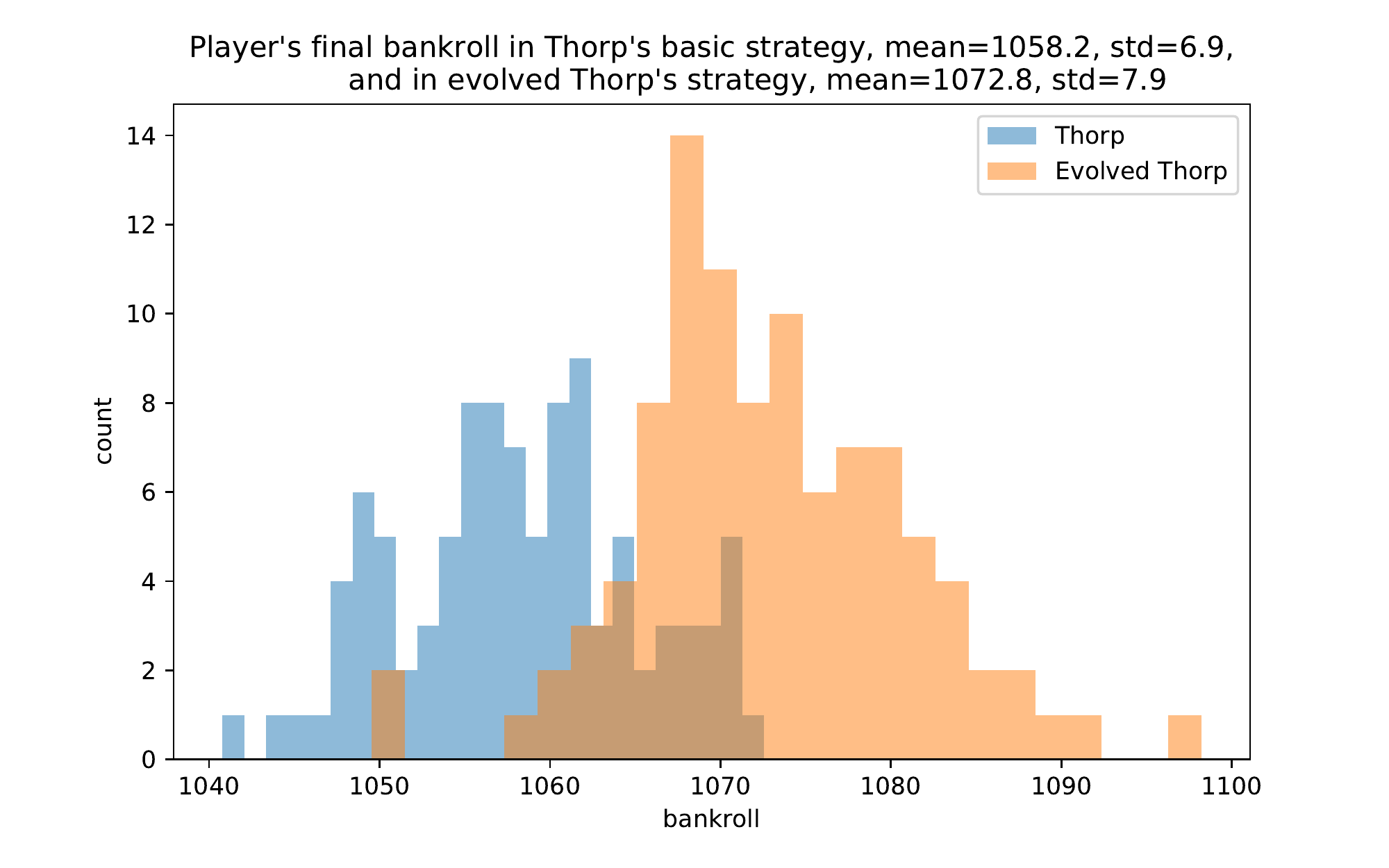}
 \caption{ \label{evolved_thorp_compare} 
 Comparison of the Thorp's basic strategy and the basic strategy obtained by evolution
 of the Thorp's basic strategy. Both histograms are obtained by running 100 simulations
 where in each simulation we play 1000 games and calculate the mean
 final bankroll (over the entire population).
 }
\end{center}
\end{figure}

In section \ref{section_evolving_random} we calculated the evolved basic strategy
by starting from a population of randomly initialized strategies.
We noticed that the resulting basic strategy, while largely similar to the Thorp's basic
strategy, has nevertheless a few differences from it.
The money performance of the Thorp's and the evolved basic strategy is quite similar,
when evaluated by playing $N=10^4$ games of $b=\$2$ bet per game.
This suggests that deviations in a certain vicinity of the Thorp's basic strategy are not very strongly
penalized, when the number of games played to evaluate the strategy fit (and the size of the bets) is not very large.
On the other hand, as we have shown in section \ref{section_evolving_random}, when evaluated by playing
$N=10^5$ games, the Thorp's strategy shows a statistically significant advantage over the evolved strategy.

Therefore in order to optimize performance of the evolved strategy, we need to use a larger number of
games to evaluate a strategy in the population. This will substantially increase the amount
of time required for an evolutionary optimization. To circumvent this issue we can start evolving
from the Thorp's basic strategy, and 
address the question of its evolutionary stability. 

To this end we initialize population of the Thorp's strategies \ref{thorp_split_soft_hard_dd}, \ref{thorp_stand} and let it evolve for $\tau=1000$
rounds. All the evolutionary parameters are chosen to be exactly the same as in section \ref{section_evolving_random},
in particular we will be using $N=10^4$ rounds of play of $b=\$2$ bet each to rank the strategies.

In figure \ref{mean_fit_random_thorp} we plot time dependence of the mean fit score (\ref{mean_score_most_fit})
of the population. We notice that performance improves, transitioning between the two (meta-)stable values, due to a spreading favorable mutation. This is to be contrasted with the
gradual evolution of the initially random strategies, shown in figure \ref{mean_fit_random}.
This is because there are many ways to start optimizing the originally
random strategy. While the Thorp's basic strategy is already very well optimized, and it
takes some time to find and spread those rare mutations which can optimize it even further.
The later steps of the random strategies evolution in figure \ref{mean_fit_random},
when zoomed in, also appear like evolution shown in figure  \ref{mean_fit_random_thorp}.

The result of running 1000 rounds of evolution is presented in figures 
\ref{mean_split_evolved_thorp}, \ref{mean_soft_double_down_evolved_thorp},
\ref{mean_hard_double_down_evolved_thorp},
\ref{mean_soft_stand_evolved_thorp},
\ref{mean_hard_stand_evolved_thorp}, where we provide de-serialized mean strategy chromosome for $\alpha M=250$
of the most fit chromosomes. As in section \ref{section_evolving_random} we used the fit
scores as the weights to calculate the mean strategy.

As in section \ref{section_evolving_random}
we can choose a specific strategy from the mean strategy, by setting a gene to be $1$
if the corresponding mean gene is larger than $a=0.95$, and setting it to $0$ otherwise.
There are a very few distinctions between the resulting evolved Thorp's strategy and the
original Thorp's strategy \ref{thorp_split_soft_hard_dd}, \ref{thorp_stand}.
Contrasted with the original Thorp's strategy, the evolved strategy does not
recommend to split
(6,6) against dealer's 6 and
(7,7) against dealer's 8.
Evolved Thorp's strategy also recommends to 
double down on (A,7) against dealer's 2, and does not recommend
to double down on hard 11 against dealer's ace.
Among inessential mutations, the mean (6,5) gene in the soft stand matrix being equal to 0.98, has no phenotypic
expression, because the soft count starts with 12.

Performance of the resulting strategy is plotted in figures 
\ref{evolved_thorp_player_bankroll}, \ref{evolved_thorp_split_dd_win_loss}.
From figure \ref{evolved_thorp_player_bankroll} we conclude that the
mean player's bankroll $\$ 1065.6$ and the standard deviation $\$234. 6$ imply
the $95\%$ confidence interval $[\$ 1051.0, \$ 1080.2]$ for the player's bankroll,
and $[ 0.23\%,0.36\%]$ for the player's edge.

We can perform a $t$-test quantifying the difference in performance of the Thorp's
basic strategy and the evolved Thorp's basic strategy, similarly to (\ref{t_mean_difference}),
giving about $83\%$ confidence that the mean values of the player's final bankroll
in the Thorp's and the evolved Thorp's strategy tests come from distributions with different means.
This does not seem to be conclusive enough that the evolved Thorp's strategy has a superior 
performance to the Thorp's strategy.
In figure \ref{evolved_thorp_compare} we compare performance of the Thorp's basic strategy
and the evolved basic strategy. This is analogous to our earlier comparison of 100 mean bankrolls, see figure 
\ref{evolved_thorp_performance}, between the basic strategy evolved from the random
population in section \ref{section_evolving_random}, and the Thorp's basic strategy.

In order to obtain a more significant statistical test of the difference between
the Thorp's and the evolved Thorp's basic strategy, we run a simulation 1000
times, with $N=10^5$ rounds of game, of $b=\$2$ bet each game.
We start with the player's bankroll of $\$ 5000$.
The resulting
final's player's bankroll for the Thorp's strategy has the mean $\$5578.3$,
and standard deviation $\$ 724.9$. The mean final bankroll for the evolved Thorp's strategy is $\$ 5738.4$,
and the standard deviation is $\$ 733.4$. The corresponding $t$-value for the difference, (\ref{t_mean_difference}),
is equal to $-4.9$, indicating $99.9\%$ confidence that the evolved Thorp's strategy
has the higher mean edge than the Thorp's strategy.

\section{Conclusions and discussion}
\label{section_conclusions}

In this paper we studied systematically the evolutionary programming approach to the
selection of an optimal basic strategy for playing blackjack. 
We demonstrated that under the specific blackjack rules chosen in this paper the basic strategy for one full
deck, proposed in \cite{Thorp1962}, has about 0.26\% edge in favor of the player. We suggest that our methodology can be applied to other
variations of the blackjack rules, analogously to \cite{Fogel2004}.

We demonstrated that evolutionary selection can quickly (in about 100 generations)
saturate to an optimal strategy which performs similarly well to known basic strategies
in the literature, such as the Thorp's basic strategy.
We have addressed the question
of the distinctions between the Thorp's basic strategy and the evolved basic strategy,
and suggested that there's an entire subset of parameters with the corresponding
basic strategies having apparently a similar performance,
when the number of rounds played and the bet size are taken small enough. The difference in performance
becomes visible only when the number of rounds of play/size of bet are scaled up.
We confirmed this statement explicitly, by increasing the number of rounds played
to evaluate the strategy performance from $10^4$ to $10^5$.
This resulted in a statistically significant difference between the strategies.

To further test our approach we have looked at evolution of the population of strategies
originally initialized to the Thorp's basic strategy. We observed that in the considered setup it took about 500 evolutionary steps
to create and spread a favorable mutation, which improves performance of the population.
The improvement has been shown to be statistically insignificant, when tested on $10^4$
rounds of play, and statistically significant, when tested on $10^5$ rounds of play.
The evolved Thorp's strategy gives the player about 0.33\% edge.
It would be interesting to extend
the approach discussed in this paper to the framework which would allow to take into account 
a specific composition of the player's hand, further refining the basic strategy.

The methods proposed in this paper can be used to calculate optimal
blackjack strategies for the deck with removed ranks. Such an approach
is an evolutionary counterpart of the simulation method originally used
by Thorp to determine the effect of removal of some cards from the deck
on the edge of the player, who uses the corresponding optimal strategy, see \cite{Thorp1962}.
We have performed this analysis elsewhere and found that evolutionary selection
acting on the population of strategies initialized randomly
optimizes the best strategy in a few hundreds of evolutionary steps, adapting to the conditions of the
deck with removed cards. Our results for the player's edge in games with deck with removed ranks agree
well with the results of \cite{Thorp1962}.

The results in this paper have been generated using the code written by the author
in C++ and Python. The original code is published on the author's GitHub page,
and can also be requested by contacting the author. On MacBook
Pro 3.3GHz Intel Core i7 (2016) it took 
about 100 seconds to calculate one generation of evolution (5000 games) in sections \ref{section_evolving_random},
\ref{section_evolving_thorp}, and about 20 seconds to test a typical strategy for 1000 games with about 10200 rounds
each in sections \ref{thorp_section}, \ref{section_evolving_random},
\ref{section_evolving_thorp}.

\small

\section*{Acknowledgments}

We are grateful to B.~Galilo for discussions, and A.~Teimouri for discussions and comments on the draft of this paper.


\section*{References}



\begin{thebibliography}{10}



 
 \bibitem{Baldwin1956}{R.~R.~Baldwin, W.~E.~Cantey, H.~Maisel, and J.~P.~McDermott,
 The Optimum Strategy in Blackjack,
 \textit{J. Am. Statist. Assoc.}, 51, 275, 429-439, 1956.}
 
 
 \bibitem{Thorp1961}{E.~O.~Thorp, 
 A favorable Strategy for Twenty-One,
 \textit{Proc. Natl. Acad. Sci.}, 47(1), 110-112, 1961.}
 
  \bibitem{Thorp1962}{E.~O.~Thorp, 
 Beat the Dealer: A Winning Strategy for the Game of Twenty-One,
 \textit{Random House, New York}, 1962.}
 
    \bibitem{Millman1983}{M.~Millman, 
 A Statistical Analysis of Casino Blackjack,
 \textit{The American Mathematical Monthly}, 90, 7, 431-436, 1983.} 
 
 
    \bibitem{Kendall2003}{G.~Kendall, C.~Smith,
The Evolution of Blackjack Strategies,
 \textit{Evolutionary Computation}, 4, 2474-2481, 2003.} 
 
  \bibitem{Fogel2004}{D.~B.~Fogel, 
 Evolving Strategies in Blackjack,
 \textit{Evolutionary Computation}, 2, 1427-1434, 2004.} 
 
  \bibitem{Yakowitz1992}{S.~Yakowitz and M.~Kollier, 
 Machine learning for optimal blackjack counting strategies,
 \textit{Journal of Statistical Planning and Inference}, 33, 295-309, 1992.} 
 
 \bibitem{Coleman2004}{R.~Coleman and M.~Johnson, 
 Genetic Algorithm-Induced Optimal Blackjack Strategies in Noisy Settings,
 \textit{Advances in Artificial Intelligence}, 467-474, 2004.} 
 



 
 
\end{thebibliography}
\end{document}